\documentclass[a4paper,10pt]{IEEEtran}
\newcommand{\attilio}[1]{\textcolor{red}{AF: #1}}

\usepackage{authblk}  
\usepackage{array}
\usepackage{graphicx} 
\usepackage{multirow}  
\usepackage{amsmath}
\usepackage{multirow} 
\usepackage{amssymb}
\DeclareUnicodeCharacter{2212}{-}
\usepackage[utf8]{inputenc} 
\DeclareUnicodeCharacter{0301}{\'{e}} 
\usepackage[T1]{fontenc} 
\usepackage{float}
\usepackage{caption}
\usepackage{booktabs}
\usepackage{array} 
\usepackage{tabularx}
\usepackage{comment}
\usepackage{xcolor}
\usepackage{hyperref}
\usepackage{tikz}

\newcommand{\setCommonTableSettings}{
    \small 
    \renewcommand{\arraystretch}{1.2} 
    \setlength{\tabcolsep}{10pt} 
}

\definecolor{lime}{HTML}{A6CE39}
\DeclareRobustCommand{\orcidicon}{%
    \begin{tikzpicture}
    \draw[lime, fill=lime] (0,0) 
    circle [radius=0.16] 
    node[white] {{\fontfamily{qag}\selectfont \tiny ID}};
    \draw[white, fill=white] (-0.0625,0.095) 
    circle [radius=0.007];
    \end{tikzpicture}
    \hspace{-1mm}
}

\newcommand{\orcidauthorA}{0000-0003-3148-3765} 
\newcommand{\orcidauthorC}{0000-0002-8337-079X}
\newcommand{\orcidauthorD}{0000-0002-9991-6822}
\newcommand{\orcidauthorE}{0000-0001-6731-3755}
\newcommand{\orcidauthorF}{0000-0003-4666-6847}
\newcommand{\orcidauthorG}{0000-0003-4274-8298}

\newcommand{\orcid}[1]{\href{https://orcid.org/#1}{\orcidicon}\hspace{1mm}}

\title{Lightweight FPGA Deployment of Learned Image Compression with Knowledge Distillation and Hybrid Quantization}

\author[1]{Mazouz Alaa Eddine \orcid{\orcidauthorA}}
\author[1]{Sumanta Chaudhuri \orcid{\orcidauthorC}}
\author[3,1]{Attilio Fiandrotti \orcid{\orcidauthorD}}
\author[1]{Marco Cagnazzo \orcid{\orcidauthorE}}
\author[2]{Mihai Mitrea \orcid{\orcidauthorF}}
\author[1]{Enzo Tartaglione \orcid{\orcidauthorG}}

\affil[1]{LTCI, Télécom Paris, Institut Polytechnique de Paris, France}
\affil[2]{SAMOVAR, Télécom SudParis, Institut Polytechnique de Paris, France}
\affil[3]{Università di Torino, Italy}

\affil[ ]{Email: alaa.mazouz@telecom-paris.fr}

\date{March 2025}

\begin{document}

\maketitle

\begin{abstract}

Learnable Image Compression (LIC) has shown the potential to outperform standardized video codecs in RD efficiency, prompting the research for hardware-friendly implementations.
Most existing LIC hardware implementations prioritize latency to RD-efficiency and through an extensive exploration of the hardware design space.
We present a novel design paradigm where the burden of tuning the design for a specific hardware platform is shifted towards model dimensioning and without compromising on RD-efficiency.
First, we design a framework for distilling a leaner student LIC model from a reference teacher: by tuning a single model hyperparameters, we can meet the constraints of different hardware platforms without a complex hardware design exploration.
Second, we propose a hardware-friendly implementation of the Generalized Divisive Normalization - GDN activation that preserves RD efficiency even post parameter quantization.
Third, we design a pipelined FPGA configuration which takes full advantage of available FPGA resources by leveraging parallel processing and optimizing resource allocation.
Our experiments with a state of the art LIC model show that we outperform all existing FPGA implementations while performing very close to the original model.
\end{abstract}

\footnote{This work is carried out under the framework of the NewEmma (Neural nEtwork Watermarking for Energy efficient Mobile Multimedia Applications) project funded by DIGICOSME.}

\section{Introduction}


Learned Image Compression (LIC) \cite{IEEE_CS1, IEEE_LIC} replaces traditional coding tools in standardized codecs with an end-to-end learnable autoencoder-based model. First, the image is projected into a low-dimensional latent space by the encoder. This representation is then quantized and entropy-coded \cite{IEEE_entropy, IEEE_entropy2} into a binary bitstream. At the receiver end, the bitstream is entropy-decoded, and the decoder recovers an approximation of the original image from the latent representation.
While adding uniform noise during training helps cope with the non-differentiability of quantization, the Generalized Divisive Normalization (GDN) activation minimizes mutual information in the latent space, improving RD efficiency. 
To date, LIC outperforms early still-image standards such as JPEG and JPEG2000 \cite{IEEE_JPEG} and performs on par with the intra profile of recent standards such as H.266/VVC \cite{IEEE_CS5}.

While research on LIC largely focused on enhancing RD efficiency through improved architectures \cite{balle_end--end_2017, balle_variational_2018, minnen_joint_2018,chen_end--end_2021,IEEE_CS2}, quantization-entropy coding stages \cite{lee_context-adaptive_2018, cheng_learned_2020, balle2018efficient,IEEE_CS3}, the implications of hardware implementations are frequently overlooked. In low-power, embedded/mobile devices, hardware accelerators (FPGA/ASIC) are crucial for meeting latency targets in Frames Per Second (FPS) while meeting power and complexity constraints. Addressing real-time latency demands involves optimizing computational \cite{ODE-ODR} and power efficiency \cite{fowers_g_nodate,gan_energy-efficient_2016,choi_trainware_nodate}, ensuring cross-platform support \cite{mazouz_automated_2021}, and compatibility with edge computing devices \cite{choi_trainware_nodate} \cite{mazouz_adaptive_2019}. Hardware-aware design strategies, such as model pruning \cite{cheng_survey_2023, tan_dropnet_2020}, and fixed-point parameter quantization \cite{gholami_survey_2022}, are essential tools for bridging the gap between algorithmic design and hardware implementation in any applicative domain.

State of the art hardware LIC implementations like \cite{mazouz2025securityrealtimefpgaintegration}, \cite{jia_fpx-nic_2022, sun_real-time_2022,fpgachen2024, sun2024fpga} are highly hardware-efficient and can achieve real-time latency (in the case of \cite{sun2024fpga,fpgachen2024}) by leveraging advanced design techniques and optimizing DSP efficiency, although \cite{fpgachen2024} uses the old \cite{balle_end--end_2017} model resulting in a considerably lower RD performance compared to \cite{sun2024fpga} and \cite{jia_fpx-nic_2022}. The Currently \cite{sun2024fpga} provides the best system level and RD performance with real-time encoding at reasonable RD quality, however such design techniques imply a deep understanding of the specific hardware platform the implementation is designed for and a time-consuming exploration of the design space to meet a latency deadline.
Also, real-time operations are often achieved at the expense of RD performance, resulting in significant drops when the hardware implementation is compared to the original LIC model it is based upon.
Such drawbacks have prompted our research, where we aim to balance hardware and software optimization adopting design choices at all levels on one hand and ensuring both real-time hardware operations do not come at the expense of RD-efficiency by designing some ad-hoc components on the other.


The main contributions of this work are:

\begin{itemize}

\item Latency Model: We analytically model the relationship between convolutional channels in a LIC architecture and its latency (FPS). By showing that computational complexity dominates this relationship, we derive an upper bound for latency, enabling the dimensioning of convolutional layer width to meet target frame rates.

\item Knowledge Distillation: We introduce a framework for distilling a reference LIC teacher model into a smaller student model, while maintaining RD-efficiency. To our knowledge, this is the first work leveraging knowledge distillation for training smaller LIC models suitable for embedded deployment.

\item GDN Implementation: We present a hardware-friendly implementation of the Generalized Divisive Normalization (GDN) layer that preserves RD-efficiency on FPGA. Unlike Sun et al. \cite{sun2024fpga,fpgachen2024,jia_fpx-nic_2022}, who replace GDN with ReLU for latency improvement, our approach retains GDN's contribution to RD performance.

\item Hybrid QAT and Pruning: We propose a hybrid QAT method that outperforms Post-Training Quantization (PTQ) in RD efficiency. Coupled with mixed precision for GDN layers, this enhances RD performance while maintaining hardware efficiency. Additionally, structured pruning optimizes model size and latency, improving FPGA resource utilization.

\item Pipelined Architecture: We present a fully pipelined architecture that maximizes hardware resource utilization for through parallel patching of high resolutions inputs.

\end{itemize}

Our experiments show state-of-the-art results in terms of latency, energy efficiency and resource utilization on the ZCU102 FPGA board (Fig.\ref{fig:main_compa} and Table \ref{tab:SOTA_energy_fps}). 
We also outperform state-of-the-art competitors in RD-efficiency, achieving higher video quality than other hardware LIC methods while maintaining lower latency.
Most important, our approach allows meeting different target frame rates without a complex exploration of the hardware design space.

The outline of this paper is as follows:
Section~\ref{sec:background_related} provides background on LIC and reviews the limitations of existing hardware implementations.
Section~\ref{sec:proposed} presents our design methodology, including knowledge distillation, pruning, quantization, and hardware pipelining.
Finally, Section 4 presents experimental results for hardware-implemented models on the ZCU102 boards, covering RD efficiency, latency, power consumption, and resource utilization.
An ablation study analyzes the contribution of each key component to the overall performance gains.

\begin{figure}[htbp]
  \centering
  \includegraphics[width=0.6\columnwidth]{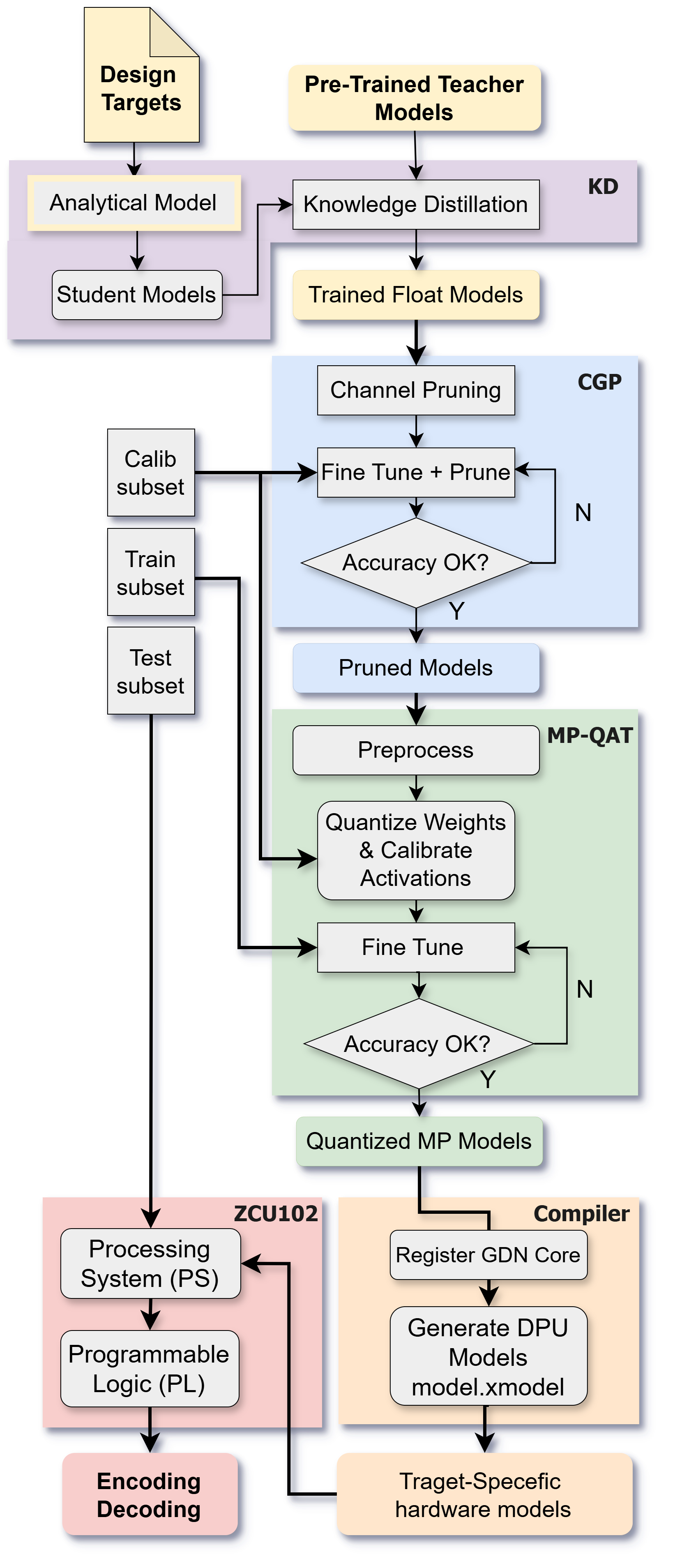}  
  \caption{The proposed workflow for training, optimizing, distilling, and deploying LIC models on hardware is comprehensive yet abstracts hardware-specific compilation through Xilinx VITIS-AI APIs}
  \label{fig:workflow_diagram}
\end{figure}
\section{Background and Related Works}
\label{sec:bacground_related}

This section first covers the background on Learned Image Compression (LIC) and then reviews existing hardware-friendly LIC approaches, highlighting the limitations that motivate our work.

\subsection{A primer on Learned Image Compression}

Learned Image Compression (LIC) has shown the ability to match or surpass standardized codecs in rate-distortion (RD) performance. Early works by Ballé et al. \cite{balle_end--end_2017} and Theis et al. \cite{theis_lossy_2017} used simpler convolutional autoencoders.
At the transmitter, a convolutional encoder $g_a$ compresses the image $x$ into a low-dimensional latent space $y$ (\textit{nonlinear analysis} coding). The latent elements are quantized with uniform noise added during training to simulate rounding, acting as a proxy for non-differentiable quantization. The quantized space $\tilde{y}$ is entropy-coded and transmitted as a binary bitstream.
In early works, each element $\tilde{y}_i$ of the latent space is learned with a non-parametric, fully factorized approach, without assumptions about the underlying distribution shape.
At the receiver, the bitstream is decoded, and a convolutional decoder $g_s$ reconstructs the approximation $\tilde{x}$ of the original image (\textit{nonlinear synthesis}).
\\
The model is optimized end-to-end by backpropagating the error gradient to minimize the cost function:

\[
L = R + \lambda \cdot D,
\]

where \(\lambda\) controls the trade-off between the rate \(R\) of the compressed bitstream and the distortion \(D\) of the reconstructed image. The rate \(R\) is estimated by an auxiliary neural network, while distortion \(D\) is measured using either Mean Squared Error (MSE) or Structural Similarity Index (SSIM).
The encoder uses Generalized Divisive Normalization (GDN) \cite{balle2015density} activation, with inverse GDN at the decoder. GDN Gaussianizes the latent space \(y\), reducing mutual information and improving compression efficiency compared to ReLU-like activations \cite{balle2018efficient}. This approach outperforms JPEG and JPEG2000 standards, despite using a relatively simple entropy model.

The above scheme was improved in~\cite{balle_variational_2018} introducing a secondary latent space known as the \textit{hyperprior}.
Rather than learning the distributions of the latents $y$ at training time, they are now predicted for each encoded image exploiting the fact that the scale $\theta$ of neighbor components usually varies jointly.
The hyperprior latent representation $z$ is first quantized using a factorized entropy model and delivered to the receiver within the bitstream.
The main quantized latent space $\tilde{y}$ is then entropy coded assuming a Gaussian distribution with $\mu = 0$ and using the scales recovered by the hyperprior decoder available also to the receiver.
The model can still be trained end-to-end yet simply adding to the cost function the estimated entropy of the hyperprior.
Despite the burden of delivering the compressed hyperprior as side information, such scheme performs on par with the H.265/HEVC Intra profile (BPG).
Clearly, this scheme is more complex due to the presence of hyper encoder and decoder, that are in turn implemented as a nested convolutional autoencoder with GDN normalization.
\begin{figure*}[htbp]
  \centering
  \includegraphics[width=0.8\textwidth]{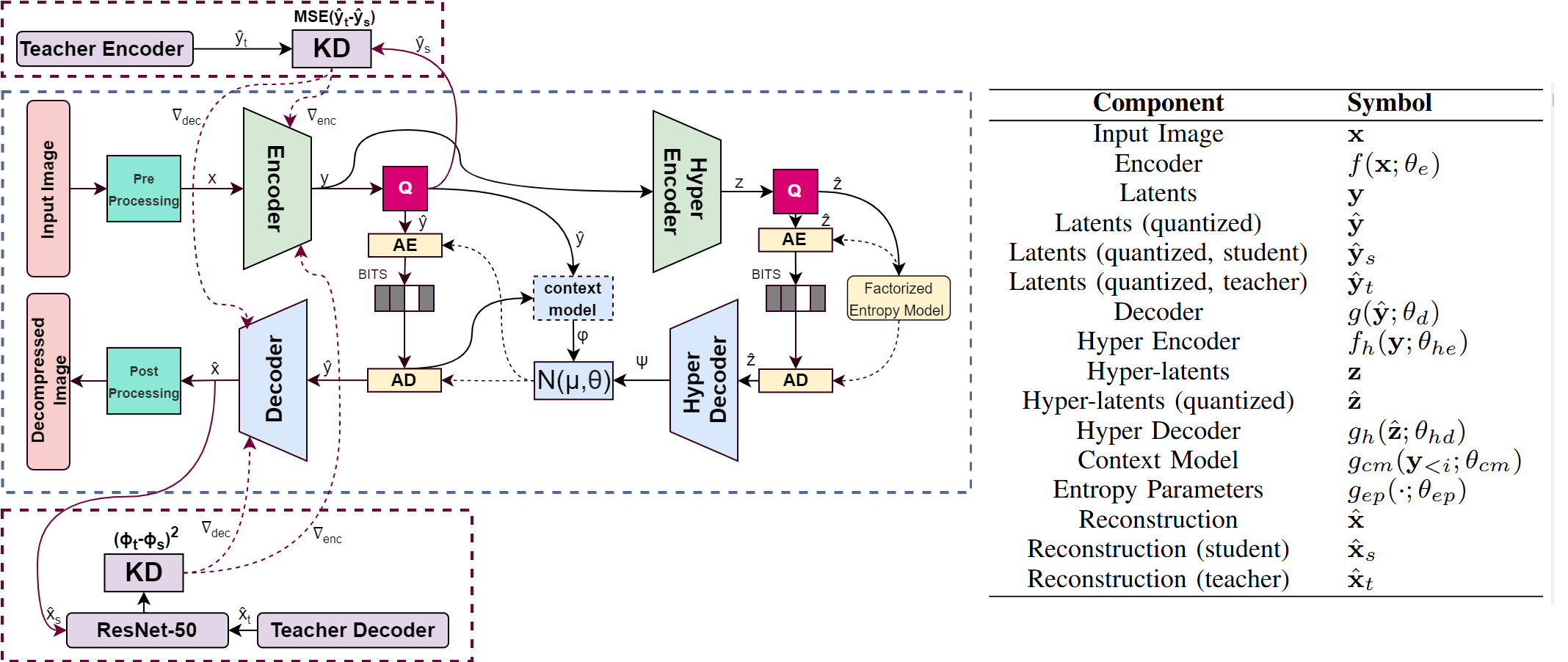}  
    \caption{Hyper-prior models capture broader image features, while context entropy model looks at the already decoded neighboring pixels (the causal context) and predicts the distribution of the next pixel based on that context
}
    \label{fig:fig:KD_diagram}
\end{figure*}

\cite{minnen2018joint,lee_context-adaptive_2018}~combined previous architectures with a context-based auto-regressive entropy model to capture local spatial correlations by exploiting already decoded image parts, An overview of this approach is shown in Fig. \ref{fig:fig:KD_diagram}. While this improves rate performance, it introduces significant latency due to sequential decoding. The model requires conditioning each pixel on previously decoded ones, limiting parallelism, which is crucial for high throughput on hardware accelerators like FPGAs. Achieving real-time compression would require architectural modifications, such as pipelining or parallel decoding techniques, to reduce latency, as shown in \cite{jia_fpx-nic_2022}.   

In~\cite{zou2022devil} local attention is exploited to combine the local-aware attention with the global-related feature learning and to the present date is among the best-performing architectures.

We note that recent hardware-friendly implementations of Learned Image Compression (LIC) have primarily focused on CNN-based models due to the challenges of implementing transformer modules in hardware. Among these, \cite{minnen2018joint} has been widely explored for its strong rate-distortion (RD) performance and hardware efficiency, which is why our work focuses on this model rather than more modern alternatives. We review notable implementations, such as \cite{fpgachen2024} and \cite{sun2024fpga}, which achieve real-time operations, but at the cost of RD performance, particularly due to the lack of Generalized Divisive Normalization (GDN) or the use of an older architecture. This gap in hardware solutions that balance both RD performance and hardware friendliness motivates our work.
The literature on hardware-friendly image compression (LIC) models is sparse, with existing works predominantly focusing on CNN-based approaches rather than Transformer-based models. This is largely due to the inherent challenges of efficiently implementing Transformer modules in hardware. Among CNN-based models, \cite{sun2024fpga} stands out for building upon \cite{sun_real-time_2022} and achieving an effective balance between rate-distortion (RD) performance and hardware efficiency, making it the current state of the art in the field. Consequently, our work adopts this model rather than more modern architectures. Notable hardware implementations, such as \cite{sun2024fpga} and \cite{sun_real-time_2022}, have demonstrated real-time processing capabilities but often at the cost of RD performance. These limitations stem from the absence of Generalized Divisive Normalization (GDN) layers and reliance on computationally intensive architectures from the literature. Additionally, \cite{jia_fpx-nic_2022} targets high-resolution deployment, achieving 4K video decoding, though at frame rates significantly below real-time requirements. The current lack of a hardware solution for LIC that simultaneously delivers hardware efficiency and superior RD performance underscores the motivation for our work, which seeks to develop novel techniques to effectively train and deploy computationally intensive LIC models for real-time applications.

\section{Proposed Method}
\label{sec:proposed}

In this section, we present our proposed method, which begins with knowledge distillation from a large teacher model to a smaller student. The student model is then implemented using a hardware-friendly GDN/iGDN, followed by pruning, mixed-precision quantization, and final deployment on hardware.

\subsection{Knowledge Distillation for LIC}
\label{sec:proposed_kd}
This section proposes a novel Knowledge Distillation (KD) scheme to transfer knowledge from a larger reference \emph{teacher} model to a smaller \textit{student} LIC model. We first outline a strategy to design a student model with lower complexity than the teacher, followed by the procedure to distill knowledge from the teacher into the student.

\subsubsection{Designing and training a student model}

We recall that, given a reference LIC model, our ultimate goal is to deploy a reference LIC model on a hardware platform while meeting a target latency. In approaches like \cite{sun2024fpga}, model complexity is reduced by altering the topology (e.g., replacing GDN activations with ReLUs) and extensively exploring the hardware design space. However, modifying the model topology often leads to a significant loss in RD efficiency, while exploring the hardware design space can be time-consuming, resource-intensive, and highly demanding in terms of hardware expertise.
In models like \cite{balle_variational_2018}, the main source of complexity stems from the filtering operations performed by the convolutional layers. Multiple hyperparameters contribute to this complexity, including the number of layers, channels per layer, filter sizes, down-sampling strategies, etc. Our approach controls model complexity by reducing the number of channels per convolutional layer while keeping the overall model topology unchanged. Exploring a single hyperparameter is far simpler than navigating the extensive hardware design space. Moreover, this approach aligns with established practices in the LIC literature, where reducing channels is a common strategy to create smaller, more computationally efficient models \cite{balle_variational_2018, balle2018efficient}.
Therefore, we begin by designing a lower-complexity student model with a fraction of the channels of the reference model. The only exception to this rule is the last convolutional layer of each encoder and decoder, where we preserve the latent space dimensions to ensure effective knowledge distillation from the teacher, as discussed below. the student model's main parameters are initialized randomly, while its hyperprior model uses the teacher's hyperprior. The student is trained by minimizing the same loss function, $\mathcal{L}_{\text{initial}} = \mathcal{R} + \lambda \mathcal{D}$, over the same training set. However, as we show experimentally later, this student model is less RD-efficient than the reference. Therefore, knowledge distillation from the teacher model into the student is necessary to guide the student to learn a more optimized RD representation.

\subsubsection{Fine-tuning the student through output and latent space knowledge distillation}

After designing and preliminarily training the student model, knowledge is distilled from the teacher to recover the lost RD-efficiency. The dotted purple boxes in Fig.\ref{fig:fig:KD_diagram} illustrates how our KD strategy integrates into the training of the reference LIC model. Specifically, the student is fine-tuned by minimizing the following loss function:
\begin{equation}
\mathcal{L}_{\text{KD}} = \alpha~\mathcal{L}_{\text{latent}} + \beta~\mathcal{L}_{\text{perc}} + \gamma~(\mathcal{R} + \lambda \mathcal{D}),
\label{eq:kd}
\end{equation}
\noindent
where the various terms are as follow.
The latent space loss term \(\mathcal{L}_{\text{latent}}\) measures the distance between the teacher's (\(\mathbf{z}_{t}\)) and the student's latent representations (\(\mathbf{z}_{s}\)) as 
$\mathcal{L}_{\text{latent}} = \text{MSE}(\mathbf{z}_{t}, \mathbf{z}_{s})$.
This loss term drives the  student to produce latent representations of the input image that match those of the teacher.
\\
The perceptual loss term \(\mathcal{L}_{\text{perc}}\) measures instead the difference between the images decoded by the teacher (\(\mathbf{x}_{t}\)) and the student (\(\mathbf{x}_{s}\)). It consists of a perceptual loss computed by comparing feature maps extracted from specific layers of a pre-trained ResNet-50 model as

\begin{equation}
\mathcal{L}_{\text{perc}} = \sum_{i \in \{3, 4, 5\}} \|\phi_i(\mathbf{x}_{t}) - \phi_i(\mathbf{x}_{s})\|^2,
\end{equation}
\noindent
where \(\phi_i(\mathbf{x})\) denotes the feature map of the reconstructed image \(\mathbf{x}\) at layer \(i\). The perceptual loss encourages the student model to focus on higher-level semantic similarities between images, rather than merely pixel-wise similarity. Layers 3, 4, and 5 of ResNet-50 are selected as they correspond to increasingly abstract feature representations, progressively moving from edges and textures to object structures. 
Namely, Layer 3 captures low-level features like edges and textures, essential for preserving fine details in compression.
Layer 4 captures mid-level features such as object parts, crucial for maintaining structural coherence.
Finally, Layer 5 captures high-level features, representing more abstract and semantic information, essential for preserving the perceptual quality of the image.
Finally, the third term in the loss function is the same RD cost term used to preliminary train the student. During fine-tuning, the three terms of the loss function loss weights $\alpha, \beta, \gamma$ are dynamically adjusted as described in \ref{sec:experimental_KD}.


\subsection{Efficient GDN design for FPGA implementation}
\label{sec:proposed_gdn}

\begin{figure}[htbp]
  \centering
  \includegraphics[width=0.8\columnwidth]{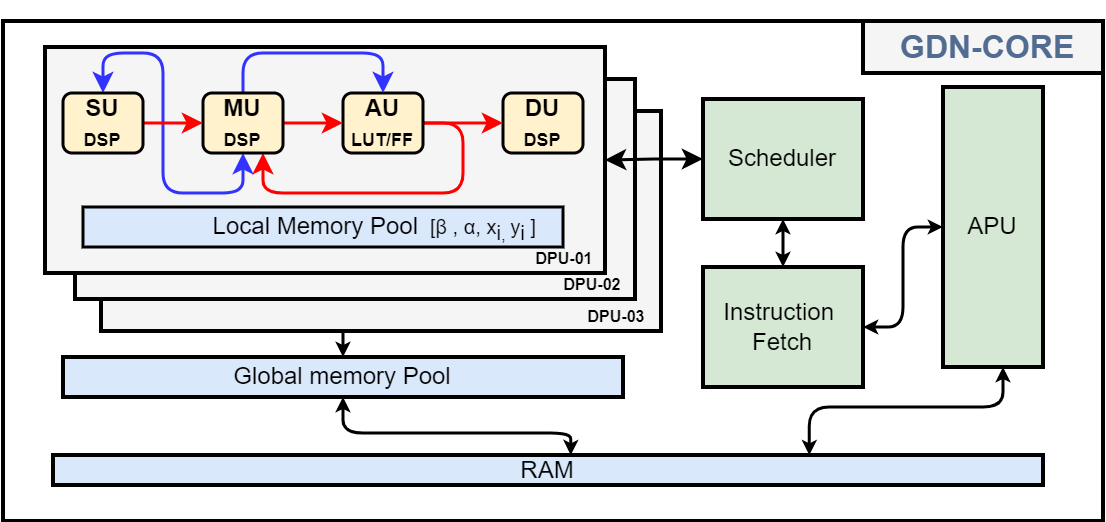}  
  \caption{Custom GDN/iGDN core integration, red for GDN, blue for iGDN pipeline using Square Unit, Multiply Unit, Add Unit and Division Unit}
  \label{fig:GDN_fpga}
\end{figure}

Other works \cite{jia_fpx-nic_2022, sun_real-time_2022, fpgachen2024,sun2024fpga} replace the GDN layer with ReLU activation to simplify the implementation. For our FPGA-based LIC architecture, we are the first to propose a custom Generalized Divisive Normalization (GDN) hardware engine to optimize RD performance. The GDN layer is essential for reducing statistical dependencies between activations in LIC models, which improves compression quality.
Given an input feature map $\mathbf{x} \in \mathbb{R}^{H \times W \times C}$, where $H$, $W$, and $C$ are the height, width, and number of channels respectively, the GDN operation is defined as:

\begin{equation}
y_i = \frac{x_i}{\left( \beta_i + \sum_j \gamma_{ij} x_j^2 \right)^\alpha}
\end{equation}

Where:
- $y_i$ is the normalized output,

- $x_i$ is the input activation from channel $i$,

- $\beta_i$ is a learned bias parameter ensuring positive output,

- $\gamma_{ij}$ is a learned parameter controlling the interaction between 
different channels,

- $\alpha$ is a parameter controlling the non-linearity, typically fixed to 0.5.

In the decoder side, the inverse GDN (iGDN) operation is used, defined as:

\begin{equation}
x_i = y_i \left( \beta_i + \sum_j \gamma_{ij} y_j^2 \right)^\alpha
\end{equation}

To efficiently implement the GDN layer on FPGA, we decomposed the mathematical operations into hardware-optimized components, leveraging the inherent parallelism and resource flexibility of the FPGA. First, the element-wise squaring operation is mapped directly onto DSP slices, which are well-suited for fast multiply-and-accumulate (MAC) operations. This minimizes latency and ensures high throughput for squaring, which is crucial given the dense computation of the GDN normalization.
The division operations, critical for computing normalized activations, are also assigned to DSP slices but are carefully pipelined to ensure a continuous flow of data. By using a reciprocal approximation technique, we reduce the complexity of the division operation, converting it into a more efficient multiplication that fits within the limited FPGA resources without compromising on accuracy.
For the weighted summation of the normalization, we utilize BRAMs (Block RAM) to store intermediate results and parameters. BRAM offers low-latency access to these values, ensuring that the summation step remains efficient even as the size of the feature maps scales up. By distributing the BRAM across multiple FPGA regions, we further enhance parallelism and enable multiple summation operations to occur concurrently across different regions of the GDN layer.
The exponentiation step, which involves computing the square root, is handled using a custom LUT-based approach. Instead of relying on traditional arithmetic units, which would significantly increase the resource usage, we implemented a piecewise-linear approximation of the square root function. This approximation is stored in LUTs, enabling rapid lookups during execution. The precision of this LUT-based approach is adjustable, allowing for dynamic trade-offs between accuracy and resource consumption based on application needs.

\begin{figure}[htbp]
  \centering
  \includegraphics[width=0.7\columnwidth]{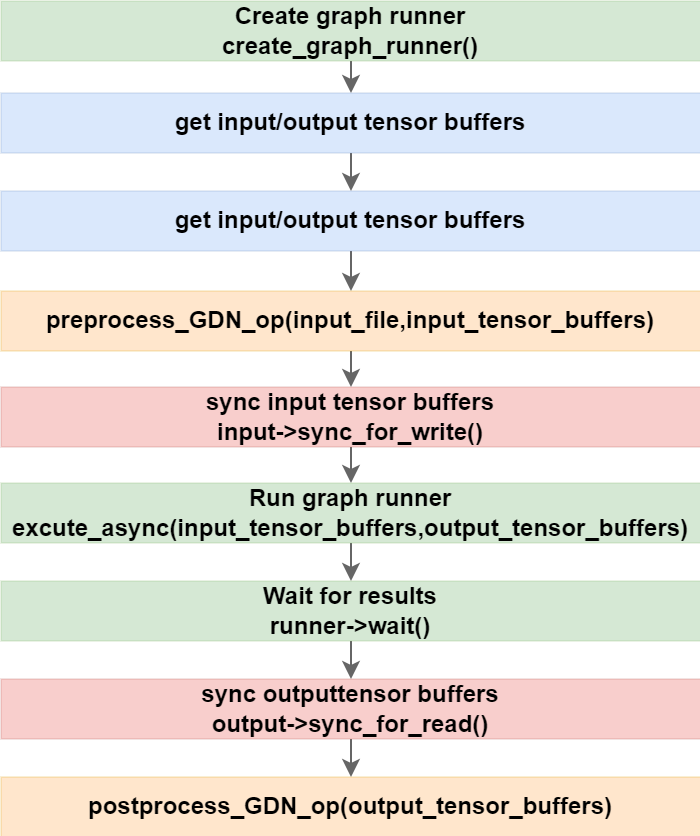}  
  \caption{Registering the custom GDN core with the XIR  
}
  \label{fig:register_gdn}
\end{figure}

Given the quantization sensitivity of the GDN layer, especially in its non-linear components, we designed a custom mixed-precision quantization strategy. While the rest of the neural network uses lower-precision 8-bit fixed-point representation for weights, activations, and biases, the GDN layer operates with 32-bit fixed-point precision for its core calculations. This ensures that the precision loss introduced by quantization does not adversely impact the rate-distortion performance of the overall model. The fixed-point operations for GDN are meticulously tuned to balance the accuracy-resource trade-off, ensuring minimal degradation in model quality while optimizing FPGA resource usage, especially with regards to DSP slices and BRAM.
The integration into the Vitis AI \cite{vitis_ai} environment required custom extensions to the Xilinx Intermediate Representation (XIR). We registered the GDN layer as a custom operation within the XIR framework, enabling the Vitis AI compiler to offload the GDN computations to a specialized Deep Processing Unit (DPU) core rather than relying on the embedded CPU. By synchronizing the GDN layer’s input-output buffers with those of the other xmodel runners, the DPU can operate with the rest of the model pipeline. This setup, as illustrated in Fig.\ref{fig:register_gdn}.

The custom GDN layer is integrated into the FPGA workflow by registering it as an operation within the Vitis AI framework, as shown in Fig.\ref{fig:register_gdn}. The process begins by creating the graph runner with \texttt{create\_graph\_runner()}, followed by allocating input and output tensor buffers via \texttt{get\_input/output\_tensor\_buffers()}. These buffers are synchronized with the FPGA memory using \texttt{input->sync\_for\_write()} before execution.
The GDN operation is performed asynchronously using \texttt{execute\_async(input\_tensor\_buffers, output\_tensor\_buffers)}, which offloads computation to the DPU. Once the operation is completed, the result buffers are synchronized back with \texttt{output->sync\_for\_read()}, and postprocessing is applied via \texttt{postprocess\_GDN\_op()}. This flow ensures efficient integration of the GDN layer, allowing it to run on the DPU in parallel with other operations, optimizing resource use and minimizing latency.

\subsection{Channel pruning}
\label{sec:proposed_pruning}

\begin{figure}[htbp]
  \centering
  \includegraphics[width=0.8\columnwidth]{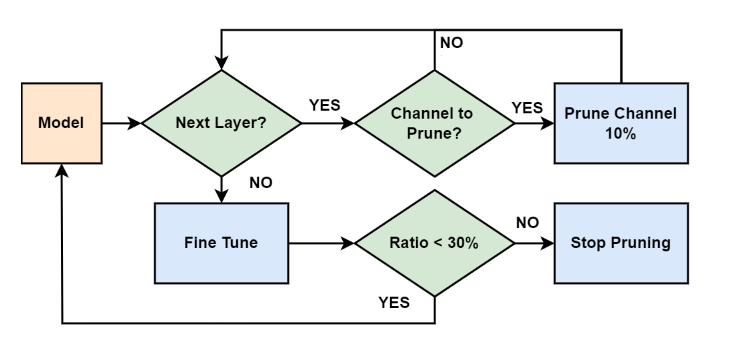}  
  \caption{The model is iteratively pruned 10\% of its filters in three iterations until 30\% sparsity is achieved, fine-tuning restoring the lost RD-efficiency.}
  \label{fig:prunning}
\end{figure}

Neural networks need to be over-parameterized to be successfully trained, so we prune the student LIC model to learn a sparse network topology and reduce FLOPs further.
While unstructured pruning (i.e., pruning single weights) has the potential to achieve higher sparsity, the DPU instructions are not designed to deal with parameter-level sparsity.
Channel pruning is instead more hardware-friendly and can be implemented with most inference architectures, most importantly with Xilinx DPU compilation instructions \cite{gao_dpacs_2023}. 
Fig.~\ref{fig:prunning} (overall diagram-blue box) details the iterative prune-and-finetune procedure. 
We shoot for pruning 30\% of the convolutional filters across three pruning iterations.
The initial input to the procedure is the student model refined as described above.
At each iteration and for each layer, 10\% of the filters with lowest L2 magnitude are pruned from the model.
If a neuron is left without filters, the entire neuron is pruned from the layer.
As pruning filters impairs the model RD-efficiency, the model is fine-tuned until the previous performance is restored.
This procedure is repeated three times until a 30\% filter level sparsity is achieved.
We found that this iterative pruning strategy enables better sparsity-performance tradeoff than single-pass pruning.
The filter pruning ratio directly affects the computational complexity as FLOPs (pruned model) = (1 – ratio) × FLOPs (original model). We will explore the the improvements gained by pruning the model, most notably the significant reduction in latency, model size and FLOPs in the experimental section.

\subsection{Mixed precision quantization}
\label{sec:proposed_quantization}

\begin{figure}[htbp]
  \centering
  \includegraphics[width=0.7\columnwidth]{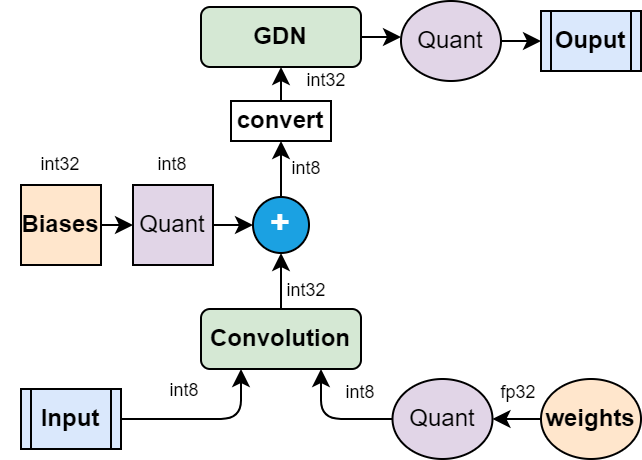}  
  \caption{Our mixed precision quantization strategy where GDN layers are quantized over 32 bits while the rest of the model is quantized over 8 bits.}
  \label{fig:quant}
\end{figure}

The parameters that survived pruning are then quantized towards integers arithmetic operations over FPGA. 
In most hardware LIC implementations \cite{sun_real-time_2022,sun2024fpga,jia_fpx-nic_2022}, all parameters are quantized to 8-bit precision to minimize latency.
Conversely, we propose a mixed-precision quantization scheme where different model components are assigned different bit numbers.
Specifically, the Generalized Divisive Normalization (GDN) layer is assigned a higher precision of 32-bit, while the rest of the model operates at 8-bit precision. This decision is motivated by the fact that the GDN layer, while only responsible for approximately 4\% of the total FLOPS in the model, plays a critical role in improving the RD efficiency~\cite{balle_integer_2018}.
We recall that GDN’s primary function is to normalize activations across channels, promoting decorrelation in the latent space and improving RD efficiency.
Lower precision (8-bit) introduces quantization noise in GDN’s operations such as division and normalization, which diminishes the accuracy of these transformations. By assigning 32-bit precision to the GDN layer, we reduce this quantization noise, leading to better normalization and decorrelation, ultimately enhancing rate-distortion performance, as we experimentally show also later on.
Fig.\ref{fig:quant} illustrates the overall quantization strategy.
Preliminarily, min and max values are extracted from the different model layers, allowing the quantizer to derive the optimal scale and zero point from the float model using the calibration subset. This subset includes 1000 random images from CLIC \cite{toderici_wenzhe_2020}, DIV2K \cite{agustsson_ntire_2017}, and Open Images \cite{kuznetsova_open_2020}, used only for a forward pass. Symmetric quantization is achieved by setting the zero point to zero, with saturation applied to minimize quantization differences.
As quantization impairs the model performance, the model undergoes a Quantization Aware Tuning (QAT) is then tuned again for 1000 iterations over a subset of 5000 images from CLIC, DIV2K, and Open Images. Simulated quantization operations are introduced on both weights and activations, except for the GDN layer, which maintains 32-bit precision to avoid losing the benefits of accurate normalization. During each forward pass, operations use the quantized weights and activations, and gradients are computed with respect to the quantized parameters using a Straight-Through Estimator (STE) \cite{jacob_quantization_2018} to approximate gradients through the quantization process. The outcome of the quantization process is an LIC model that exclusively relies on integer operations, meeting the requirements for efficient FPGA deployment.

\subsection{Design Space Exploration}
\label{sec:proposed_hardware}

\begin{figure*}[htbp]
  \centering
  \includegraphics[width=0.7\textwidth]{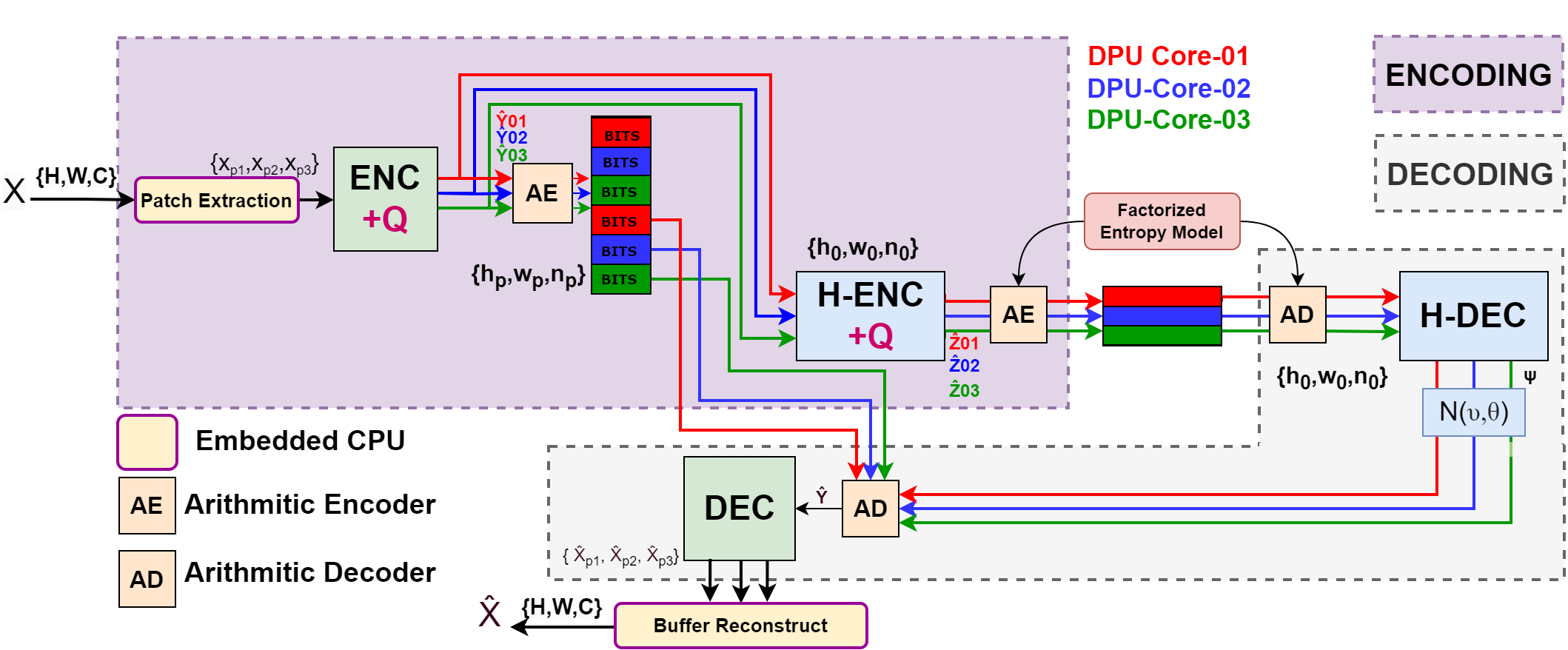}  
  \caption{Fully pipelined, patches are processed in parallel by three dedicated different computational cores, and intermediate outputs are available more quickly and passed immediately to the next module, this reduces the number of idle resources and utilizes the available bandwidth more efficiently. 
}
  \label{fig:pipe}
\end{figure*}

In this section, we finally describe the hardware configuration we use for implementing the quantized LIC model model on FPGA using the Vitis AI platform \cite{vitis_ai}.
Namely, we target the Xilinx ZCU102 FPGA that houses three DPU cores running at 300 MHz. 
We propose below a fully pipelined configuration employing the common B4096-DPU core configuration.
Such configuration provides a peak of 4096 Multiply-Accumulate (MAC) operations per cycle featuring 16 input channels, 16 output channels, and processing 8 pixels per cycle.
Namely, we seek a fully pipelined configuration where each of the three DPU cores operates in parallel, as depicted in Fig.\ref{fig:pipe}.
By dividing the image into 256 × 256 patches and processing multiple patches concurrently, the fully pipelined configuration optimizes resource utilization and minimizes end-to-end latency. Each DPU core is assigned a distinct patch, ensuring continuous processing and efficient use of computational resources.  

This setup dedicates each DPU to processing different image patches simultaneously, maximizing throughput and reducing idle times. In contrast, the patch-based pipeline in \cite{jia_fpx-nic_2022} does not fully utilize available resources by only using the cores sequentially, leading to significantly lower throughput.  
Unlike \cite{sun2024fpga, fpgachen2024}, we do not perform extensive design space exploration beyond efficient pipelining. Instead, we focus on reducing the computational load of the models, enabling us to outperform the works that rely on heavy design space exploration—without without requiring extensive hardware expertise or specialized knowledge.


\begin{figure}[t]
  \centering
  \includegraphics[width=0.7\columnwidth]{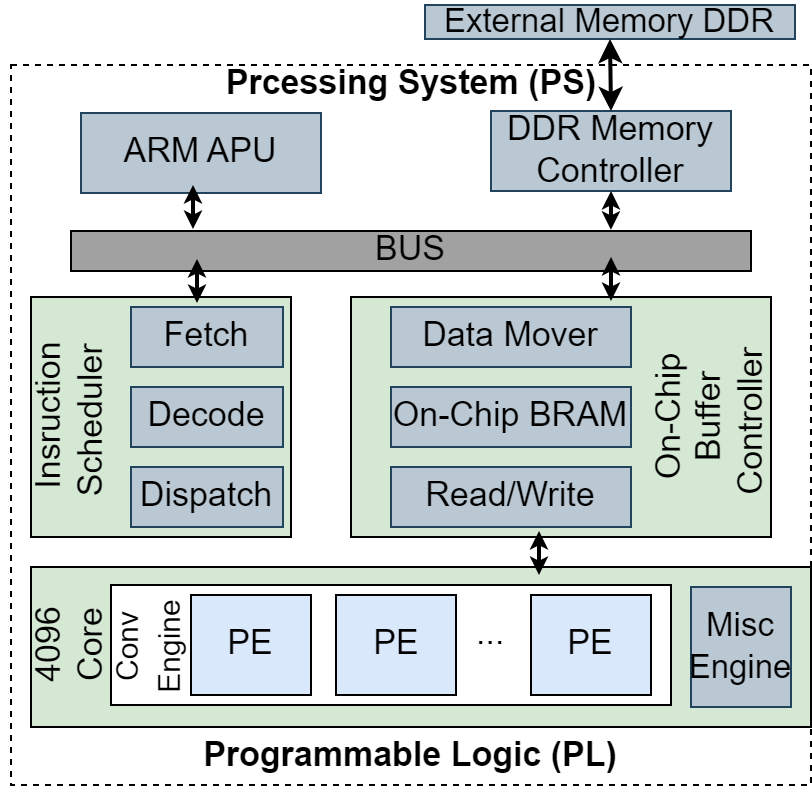}  
  \caption{Hardware architecture of the overall DPU }
  \label{fig:DPU_diagram}
\end{figure}

The model is compiled into target-specific DPU instructions in the form of an xmodel. The XIR compiler constructs an internal computation graph as an intermediate representation (IR), it then performs multiple optimizations, such as computation node fusion, and efficient instruction scheduling by exploiting inherent parallelism or exploiting data reuse on the specific target. This unified IR is broken down into subgraphs based on the corresponding control flow, for our models multiple DPU subgraphs are compiled as the custom GDN layers require separate graphs. Finally, an instruction stream is generated for each subgraph and everything is serialized into a target-specific, compiled xmodel file. This xmodel is encrypted and then deployed on the FPGA using the runtime API through the embedded processor.
We deploy the models on the Zynq-UltraScale+ MPSoC DPU architecture -DPUCZDX8L, seen Fig.\ref{fig:DPU_diagram}. The nomenclature reflects the target FPGA, quantization bit-width, and design objective, where 'L' denotes low latency.
The DPU provides multiple configurations to trade off PL resources for performance. The B4096 core configuration yields the highest throughput with a peak of 18×16×16×2=4096 MAC/cycle. Three B4096-DPUCZDX8L cores are deployed on the ZCU102. 

\subsection{Analytical Model for Estimating FPS}

This section introduces an analytical framework for estimating the Frames Per Second (FPS) of learned image compression (LIC) models deployed on the Vitis AI DPU 4096 accelerator. The framework provides a coarse-grained estimation of the model's performance before implementation, helping in the informed selection of model sizes to meet specific FPS targets. By integrating computational constraints and hardware parameters, the model offers a practical tool for hardware-aware model design.

The FPS of a model depends on the computational performance of the DPU core, as well as the model’s workload, which is determined by the number of operations required to process a frame. The FPS can be estimated by calculating the time required to process one frame using the available compute power of the DPU.

The total frame time is mainly determined by the compute time, as memory transfer time is relatively negligible given the high bandwidth available and the small size of the data being processed. Thus, the FPS can be estimated as the inverse of the compute time.

The compute time is calculated as:

\begin{equation}
T_{\text{compute}} = \frac{N_{\text{workload}}}{\text{Peak Ops per Cycle} \times \eta}
\label{eq:compute_time}
\end{equation}

Where:
\begin{itemize}
    \item \( N_{\text{workload}} \) is the total number of operations required to process one frame.
    \item Peak Ops per Cycle is the total computational power available from the DPU cores.
    \item \( \eta \) is an efficiency factor accounting for any overhead, such as layer merging and memory caching inefficiencies.
\end{itemize}

The Peak Ops per Cycle is determined by the DPU configuration, which includes the number of input and output channels per layer, pixel parallelism, and core utilization. For each convolutional layer, the total number of operations per cycle per core is given by:

\[
\text{Peak Ops per Cycle (per core)} = PP \times ICP \times OCP \times 2
\]

Where:
\begin{itemize}
    \item \( PP \) is the pixel parallelism, representing the number of pixels processed in parallel.
    \item \( ICP \) is the input channel parallelism, representing the number of input channels processed in parallel.
    \item \( OCP \) is the output channel parallelism, representing the number of output channels processed in parallel.
    \item The factor of 2 accounts for both input-output and weight operations in the convolution process.
\end{itemize}

For the DPU configuration with \( PP = 8 \), \( ICP = 16 \), and \( OCP = 16 \), the peak operations per cycle per core (denoted as \( \text{Ops}_{\text{core}} \)) is:
\[
\text{Ops}_{\text{core}} = 8 \times 16 \times 16 \times 2 = 4096
\]
With three DPU cores operating in parallel, the total peak operations per cycle (denoted as \( \text{Ops}_{\text{total}} \)) is:
\[
\text{Ops}_{\text{total}} = 3 \times 4096 = 12,288
\]
The total frame time, \( T_{\text{frame}} \), is thus:

\begin{equation}
T_{\text{frame}} = \frac{N_{\text{workload}}}{\text{Peak Ops per Cycle (total)} \times \eta}
\label{eq:frame_time}
\end{equation}

Finally, the FPS is the inverse of the total frame time:

\[
\text{FPS} = \frac{1}{T_{\text{frame}}}
\]

Substituting the expression for \( T_{\text{frame}} \), the FPS can be written as:

\begin{equation}
\text{FPS} = \frac{\text{Peak Ops per Cycle (total)} \times \eta}{N_{\text{workload}}}
\label{eq:fps}
\end{equation}

The total memory load per inference pass on a given model, \( B_{\text{load}} \), can be approximated by:
\begin{align}
B_{\text{load}} = \sum_{l=1}^L \big( & H_l W_l N_l \times \Delta_l + H_l W_l N_{l+1} \times \Delta_l \notag \\
                                     & + N_l N_{l+1} K^2 \times \Delta_l \big)
\label{eq:bandwidth_load}
\end{align}

where:
\begin{itemize}
  \item \( L \): Total number of layers in the model.
  \item \( H_l, W_l \): Height and width of the feature map in layer \( l \).
  \item \( N_l \): Number of channels (feature maps) in layer \( l \).
  \item \( N_{l+1} \): Number of channels (feature maps) in the next layer (\( l+1 \)).
  \item \( K \): Kernel size (assumes square kernels of size \( K \times K \)).
  \item \( \Delta_l \): Bit-width or precision of the parameters and activations in layer \( l \), expressed in bits.
\end{itemize}

\section{Experiments and Results}
\label{sec:experimental}

In this section we experiment with the methodology proposed in the previous section over the \cite{balle_variational_2018} LIC model.
First, we train a teacher model, we use it to teach two different students and we deploy the three models on FPGA, evaluating the different rate-distortion-complexity our method can attain against state of the art references.
Second, we ablate each element of our proposed methodology assessing its contribution to the overall performance.

\subsection{Training the Hardware Efficient Models}
\subsubsection{Knowledge Distillation}
\label{sec:experimental_KD}
To begin, we train the LIC model \cite{balle_variational_2018} from scratch to validate our experimental setup, following the original paper's methodology. Our implementation builds upon the CompressAI framework \cite{2020compressaipytorchlibraryevaluation}, which provides a standardized baseline for benchmarking.  
For transparency, we report the hardware specifications: training was conducted on an A100-PCIE-40GB GPU, using the hyperparameters presented in Table \ref{tab:hyperparameters}, ensuring reproducibility and substantiating our claim of training from scratch.  
The model is trained using a randomized train/validate/test split of the combined CLIC, DIV2K, and a subset of the Open Images dataset. Training is performed over batches of 256 × 256 randomly cropped patches, using four $\lambda$ values and the hyperparameters listed in Table~\ref{tab:hyperparameters}.  

We refer to this model as the \textit{Teacher}, which is then used to train several student models with 160 and 128 filters per layer, denoted as \textit{Student-160} and \textit{Student-128}, respectively, the knowledge distillation scheme hyperparameters are presented in Table~\ref{tab:hyperparameters} .   
Next, the teacher and the students are evaluated on the Kodak dataset following the same procedure described in related works like \cite{jia_fpx-nic_2022}.

\begin{table}[htbp]
\centering
\caption{Hyperparameters for the KD Training Scheme}
\renewcommand{\arraystretch}{1.3} 
\footnotesize 
\begin{tabular}{>{\centering\arraybackslash}p{2.8cm}>{\centering\arraybackslash}p{1.2cm}>{\centering\arraybackslash}p{1.5cm}>{\centering\arraybackslash}p{1.5cm}}
\hline
\multirow{2}{*}{\textbf{Hyperparameter}} & \multirow{2}{*}{\textbf{Initial}} & \multicolumn{2}{c}{\textbf{Fine Tuning}} \\ 
\cline{3-4}
                                     & \textbf{Training} & \textit{Early} & \textit{Late} \\ 
\hline
\text{Learning Rate}            & 1e-3            & 1e-4 & 1e-6 \\
\text{Decay [Step, Rate]}      &0.5M , 0.9      & 25k , 0.9 & 25k , 0.9 \\
\text{Batch Size [256x256]}     & 16            & 8 & 8 \\
\text{\(\alpha\)}               & NA             & 0.1 & 1.0 \\
\text{\(\beta\)}                & NA             & 1.0 & 0.1 \\
\text{\(\gamma\)}               & NA             & 0.5 & 0.5 \\
\text{Steps}                   & 5M             & 250K & 250K \\
\text{\(\lambda\)}              & \multicolumn{3}{c}{0.0016, 0.0032, 0.0075, 0.045} \\
\hline
\end{tabular}%
\label{tab:hyperparameters}
\end{table}

First, 720p (1280x720) HD images are produced repeatedly tiling the Kodak images.
The tiled image is split into overlapping 256x256 patches with 56x56 stride, each patch is independently encoded and decoded, and finally, the decoded patches are reassembled into the whole picture. While such operations introduce some overhead, they are needed for HD and above pictures because of the otherwise impractical memory and bandwidth footprint on the FPGA and the hardware engine's requirements for fixed size input.

Early in the fine-tuning process, we set $\alpha=1.0$ and $\beta=0.1$ to encourage the student latent space to match that of the teacher, thereby achieving a similar latent space entropy. As training progresses, $\alpha$ is reduced to 0.1 and $\beta$ is increased to 1.0 to promote visual quality in the output. The transition between these two phases is automated by monitoring performance plateaus; for example, once the latent space loss $\mathcal{L}_{\text{latent}}$ stabilizes, the weights are automatically adjusted to emphasize the output space. Empirically, setting each phase to 250Ksteps satisfies our requirements on the chosen dataset.

Throughout the distillation process, the weight $\gamma$ is maintained at 0.5 to balance the rate-distortion (RD) loss and prevent catastrophic forgetting of the original end-to-end compression task. Table~\ref{tab:hyperparameters} lists all the relevant hyperparameters for the distillation process. It is worth noting that the teacher's pre-trained hyperprior model is reused throughout training.

Table~\ref{tab:GFLOPS} details the number of floating point operations (GFLOPs) required at inference time for each module depicted in Fig.~\ref{fig:fig:KD_diagram}. Notably, for the main model, the distillation process reduces the teacher's filters by approximately 16\% in Student-160, effectively halving the computational complexity; a further 16\% reduction in Student-128 slashes the overall complexity by a factor of four.

The students are then fine-tuned as described in Sec.~\ref{sec:proposed_kd} to improve their RD performance.
The teacher and the two distilled students are then pruned as in Sec.~\ref{sec:proposed_pruning}, the parameters are quantized as in Sec.~\ref{sec:proposed_quantization}, the GDN is implemented as in Sec.~\ref{sec:proposed_gdn}, and the corresponding xmodels are compiled and individually ported to the ZCU102 FPGA for benchmarking.

\begin{table}[htbp]
\centering
\caption{Number of GFLOPs required to process a 720p input after structured pruning}
\renewcommand{\arraystretch}{1.2} 
\begin{tabular}{>{\centering\arraybackslash}p{2cm}>{\centering\arraybackslash}p{1.6cm}>{\centering\arraybackslash}p{1.6cm}>{\centering\arraybackslash}p{1.6cm}}
\hline
\textbf{\begin{tabular}[c]{@{}c@{}}Module\end{tabular}} & \textbf{Teacher-192} & \textbf{Student-160} & \textbf{Student-128} \\ 
\hline
\textit{ $f(\mathbf{x}; \theta_e)$ }                 & 281.8               & 106.1             & 77.4              \\ 
\textit{$g(\hat{\mathbf{y}}; \theta_d)$}                 & 300.4              & 114.3              & 75.6             \\ 
\textit{$g_{ep}(\cdot; \theta_{ep})$ }           & 0.025               & 0.0087          & 0.006               \\ 
\textit{$f_h(\mathbf{y}; \theta_{he})$}           & 179.9                    & 179.9                    & 179.9                     \\ 
\textit{$g_h(\hat{\mathbf{z}}; \theta_{hd})$}           & 224.1                    & 224.1                   & 224.1                    \\ 
\hline
\textbf{Total Main}              & 582.2            & 220.5             & 153.1               \\ 
\textbf{Total Hyper}        & 404.0              & 404.0             & 404.0               \\ 
\textbf{Total}                    & 986.2            & 624.5             & 557.1               \\ 
\hline
\end{tabular}
\label{tab:GFLOPS}
\end{table}
\subsubsection{Pruning and Quantization}
Once we have our smaller student models, we apply iterative channel pruning and mixed-precision quantization-aware training (QAT) as described in Sections~\ref{sec:proposed_pruning} and \ref{sec:proposed_quantization}. The effects of pruning and quantization are presented in Tables~\ref{tab:pruning_floats} and \ref{tab:qat_pruned_float}, respectively.

Table~\ref{tab:pruning_floats} shows that channel pruning can reduce model size by approximately 30\%  and significantly decrease inference latency (by around 40\%). It also cuts the FLOPS by roughly 30\%, although this comes with a modest drop in rate-distortion (RD) performance (e.g., a 0.25 dB decrease in PSNR for Student-160).

Subsequently, we perform QAT for 1000 iterations using a subset of the original training data. As illustrated in Table~\ref{tab:qat_pruned_float}, QAT further compresses the model (achieving around a 43\% reduction in size for Student-160) and provides moderate improvements in inference latency, while the impact on RD performance remains minimal.

\begin{table}[htbp]
\centering
\setCommonTableSettings 
\caption{Effect of channel-wise pruning compared to the reference models}
\begin{tabular}{p{3.5cm} c c c} 
\hline
\textbf{} & \textbf{160} & \textbf{128} & \textbf{96} \\ 
\hline
\textbf{$\Delta$ Model size ↓} & -33.2\% & -27.1\% & -27.1\% \\ 
\textbf{$\Delta$ Inference latency ↓} & -39.0\% & -24.3\% & -29.4\% \\ 

\textbf{$\Delta$ Avg PSNR (dB) ↑} & -0.25 & -0.28 & -0.41 \\ 
\hline
\end{tabular}
\label{tab:pruning_floats}
\end{table}

\begin{table}[htbp]
\centering
\setCommonTableSettings
\caption{Effect of Quantization Aware Training (QAT) compared to the pruned float model}
\begin{tabular}{l c c c}
\hline
\textbf{} & \textbf{160} & \textbf{128} & \textbf{96} \\ 
\hline
\textbf{$\Delta$ Model size ↓} & -42.73\% & -48.92\% & -49.01\% \\ 
\textbf{$\Delta$ Inference latency ↓} & -12.63\% & -13.18\% & -14.05\% \\ 
\textbf{$\Delta$ Avg PSNR (dB) ↑} & +0.108 & -0.095 & -0.108 \\    
\hline
\end{tabular}
\label{tab:qat_pruned_float}
\end{table}

\subsection{Analytical Model Validation}

To validate the proposed analytical framework, we apply the model to three LIC models with different computational workloads and compare the estimated FPS with the observed FPS values.

We validate the analytical model with a set of experiments, using the ZCU102 FPGA with the following specifications:

\begin{itemize}
    \item DPU frequency: 300 MHz
    \item DPU size: B4096, 3 cores in parallel, each capable of 4096 operations per cycle
    \item Memory bandwidth: 19.2 GB/s for PL and 19.2 GB/s for PS 
    \item Input dimensions: \( H = 1280 \), \( W = 720 \)
    \item Kernel size: \( K = 5 \)
\end{itemize}

The workloads (in GOP/frame) for each model for the encoding task, after pruning, are as follows:

\begin{itemize}
    \item{192K model}: → {528.2 GOP/frame} 
    \item {160K model}: → {220.2 GOP/frame} 
    \item {128K model}: → {153.0 GOP/frame} 
\end{itemize}

We also assume the following parameters:

\begin{itemize}
    \item {Peak operations per cycle (total)}: 12,288 operations per cycle (3 DPU cores, each capable of 4096 operations per cycle)
    \item {Pipeline efficiency factor} (\( \eta \)): 0.8 (Coarsely accounting for overhead of data transfer and pipeline efficiency)
\end{itemize}

We test three different model configurations with varying numbers of channels (\( N \)):
\[
N = 192, 160, 128
\]
For each model, we use equations \ref{eq:compute_time}, \ref{eq:frame_time}, \ref{eq:fps} and \ref{eq:bandwidth_load}  to compute the FPS based on the workload and the given parameters , the FLOPs for each model are computed using the equation for \( \text{FLOPs}_{\text{model}} \) based on the number of channels. The memory load is calculated using the memory traffic equation, and the FPS is estimated using the formulas described in earlier.
The final FPS values are summarized in Table~\ref{tab:ana_results}.

\begin{table}[htbp]
    \centering
    \caption{Validation of Analytical Model for Encoding-Only Pruned Models}
    \label{tab:ana_results}
    \begin{tabular}{ccccccc}
        \hline
        $N$ & $\text{GFLOPs}$ & Predicted FPS & Observed FPS & $\Delta$  FPS ↓ \\
        \hline
        192K & 582.2 & 18.6 & 15.6 & -16.13\% \\
        160K & 220.2 & 44.6 & 43.1 & -3.36\% \\
        128K & 153.0 & 64.1 & 58.7 & -8.42\% \\
        \hline
    \end{tabular}
\end{table}
The analytical model provides a reasonable estimate of FPS across different configurations, with deviations of around 12\%. It effectively captures the transition from compute-bound performance at higher \( N \) values to memory-bound performance at lower \( N \) values. For instance, at higher \( N \), performance is mainly determined by the computational resources available in the DPU cores, while at lower \( N \), memory bandwidth becomes a more significant limiting factor.
The framework allows users to select the number of channels (\( N \)) to meet specific FPS targets. For example, to achieve an FPS target of approximately 40, the model suggests \( N = 160 \), balancing computational and memory demands. For FPS values greater than 60, reducing \( N \) to 128 or lower can minimize both computational and memory requirements, while still meeting throughput goals.
However, for lower-load models, such as the 128K configuration, there is a larger discrepancy between the predicted and observed FPS. This discrepancy likely arises because, at lower \( N \), the model reaches its memory bandwidth limit, which is not fully accounted for in the analytical model. Although smaller models may theoretically improve throughput by utilizing more available computational resources, in practice, performance is constrained by memory bandwidth in such configurations. As a result, the model cannot fully exploit the available computational resources, leading to memory-bound performance.
While the analytical model offers useful FPS estimates, it simplifies key factors such as memory access patterns, cache behavior, and intra-layer parallelism. It also does not fully consider the dynamic interaction between memory bandwidth and compute resources, or factors such as resource contention and synchronization delays. These aspects require more detailed, cycle-level analysis, similar to that in \cite{mazouz_automated_2021}, which we leave for future work.
\subsection{Performance evaluation}

\begin{figure*}[htbp]
  \centering
  \includegraphics[width=0.8\textwidth]{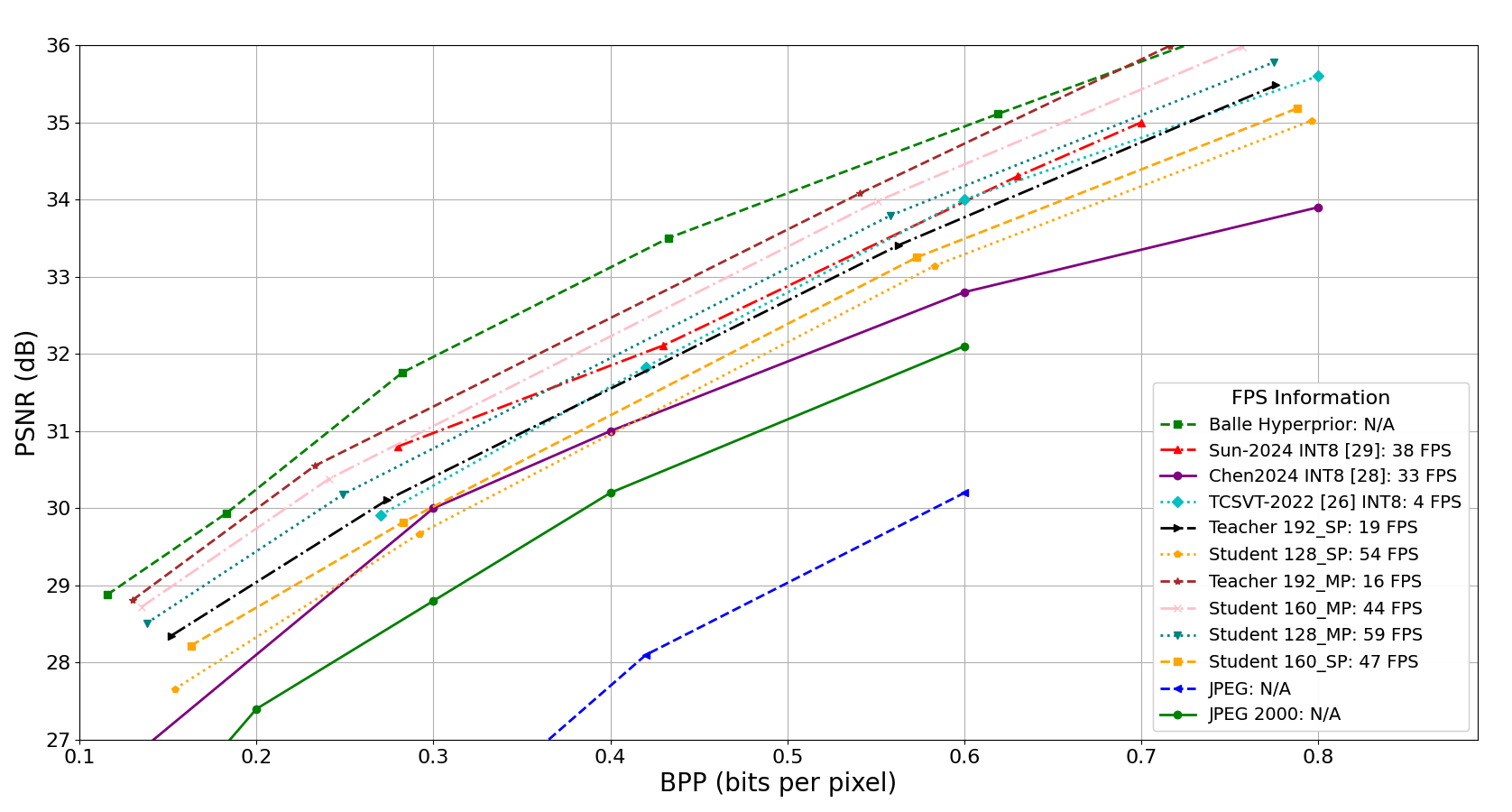}  
  \caption{Full R-D comparison with state of the art, \cite{jia_fpx-nic_2022, sun2024fpga,fpgachen2024} and standard codecs on Kodak dataset }
  \label{fig:main_compa}
\end{figure*}

\begin{table*}[htbp]
\centering
\caption{Final FPGA, BD metrics, FPS, and system-level results of various compression methods and comparisons with related works}
\setCommonTableSettings
\footnotesize
\renewcommand{\arraystretch}{1.1} 
\resizebox{\textwidth}{!}{%
\begin{tabular}{l>{\centering\arraybackslash}p{1.5cm}%
>{\centering\arraybackslash}p{2.5cm}>{\centering\arraybackslash}p{2.0cm}%
>{\centering\arraybackslash}p{2.5cm}>{\centering\arraybackslash}p{2.5cm}%
>{\centering\arraybackslash}p{2.5cm}}
\hline
\multirow{2}{*}{\textbf{Method}} & \multirow{2}{*}{\textbf{Model}} & \multirow{2}{*}{\textbf{Data Type}} & \multicolumn{3}{c}{\textbf{Performance}} & \multirow{2}{*}{\textbf{FPGA}} \\ 
\cline{4-6}
 & & & \textbf{Frame Rate} & \textbf{BD-PSNR} & \textbf{BD-Rate} &  \\
\hline
\textbf{Balle Hyperprior Baseline}  & \cite{balle_variational_2018} & fp32 & N/A & N/A & N/A & N/A \\
\hline
JPEG       & NA  & fp32 & N/A & -5.74 dB & +250.88\% & N/A \\
JPEG-2000  & NA  & fp32 & N/A & -2.99 dB & +112.60\% & N/A \\
\hline
\multicolumn{7}{l}{\textbf{Teacher Models}} \\ 
\hline
Teacher 192\_SP & \cite{balle_variational_2018} & int8 & 19 FPS & -1.33 dB & +37.63\% & ZCU102 \\
Teacher 192\_MP & \cite{balle_variational_2018} & int8/int32 & 16 FPS & \textbf{-0.66 dB} & \textbf{+17.62\%} & ZCU102 \\
\hline
\multicolumn{7}{l}{\textbf{Student Models}} \\
\hline
Student 160\_SP & \cite{balle_variational_2018} & int8 & 47 FPS & -1.68 dB & +49.35\% & ZCU102 \\
Student 128\_SP & \cite{balle_variational_2018} & int8 & 64 FPS & -1.94 dB & +57.29\% & ZCU102 \\
Student 160\_MP & \cite{balle_variational_2018} & int8/int32 & 44 FPS & -0.90 dB & +24.53\% & ZCU102 \\
Student 128\_MP & \cite{balle_variational_2018} & int8/int32 & 59 FPS & -1.23 dB & +34.57\% & ZCU102 \\
\hline
\multicolumn{7}{l}{\textbf{Related Works}} \\
\hline
Sun-2024 \cite{sun2024fpga}   & \cite{balle_variational_2018} & int8 & 38 FPS & -1.16 dB & +28.22\% & VCU118 \\
Jia-2022 \cite{jia_fpx-nic_2022}  & \cite{minnen2018joint} & int8 & 4 FPS  & \textbf{-0.51 dB} & +26.34\% & ZCU104 \\
Chen-2024 \cite{fpgachen2024}   & \cite{balle_end--end_2017} & int8 & 33 FPS & -1.54 dB & +37.12\% & ZCU102 \\
\hline
\end{tabular}%
}
\label{tab:compression_results}
\end{table*}

Fig. \ref{fig:main_compa} compares RD-efficiency (rate vs. distortion) and latency (where available) for our proposed method and multiple references benchmarked on a ZCU102 FPGA.
As an upper bound, we report the RD curve of the original LIC model \cite{balle_variational_2018} on a NVIDIA Tesla K80 GPU.
As lower bounds, we report the RD curves of the JPEG and JPEG2000 standards.
All FPGA-implemented model show a significant drop in RD efficiency mainly due to quantization: similar drops are shared by any FPGA implementation in the literature.
\\
Coming to the comparison between FPGA-based models, we take as a reference the state of the art results in \cite{sun2024fpga}.
While in the hardware LIC literature it is not unseen comparing codecs measuring the PSNR delta for some target bitrate, we will always compare pairs of RD curves through Bjontegaard metrics\cite{bjontegaard2001calculation}.
Our \textit{Teacher} model outperforms \cite{sun2024fpga} in rate-distortion (RD) efficiency, achieving \textbf{-8.95\% BD-rate savings} and \textbf{0.48 dB BD-PSNR} better image quality. However, due to its larger computational cost, it falls short of the 25 FPS target and remains below the 37 FPS achieved by \cite{sun2024fpga}.  

In contrast, \textit{Student-160} improves both RD efficiency and latency. It achieves \textbf{-5.32\% BD-rate savings} and \textbf{0.25 dB BD-PSNR} improvement over \cite{sun2024fpga}, while also surpassing its speed (43 FPS vs. 37 FPS), demonstrating the effectiveness of our proposed knowledge distillation scheme.  

\textit{Student-128}, while less RD-efficient than \cite{sun2024fpga} (\textbf{+7.80\% BD-rate} and \textbf{-0.10 dB BD-PSNR}), meets the \textbf{50 FPS real-time threshold} and is close to the \textbf{60 FPS target}, making it a viable option for latency-critical applications.  

On the other hand, \cite{jia_fpx-nic_2022} and \cite{fpgachen2024} fail to meet the required standards for different reasons. \cite{jia_fpx-nic_2022} employs a sophisticated model based on \cite{minnen2020channelwiseautoregressiveentropymodels}, achieving strong RD performance but at the cost of extremely low speed (only 4 FPS). In contrast, \cite{fpgachen2024} prioritizes speed, reaching 33 FPS, but relies on an outdated model \cite{balle_end--end_2017}, resulting in uncompetitive RD performance, for instance it under performs even our smallest student model.

\begin{figure}[htbp]
  \centering
  \includegraphics[width=1\columnwidth]{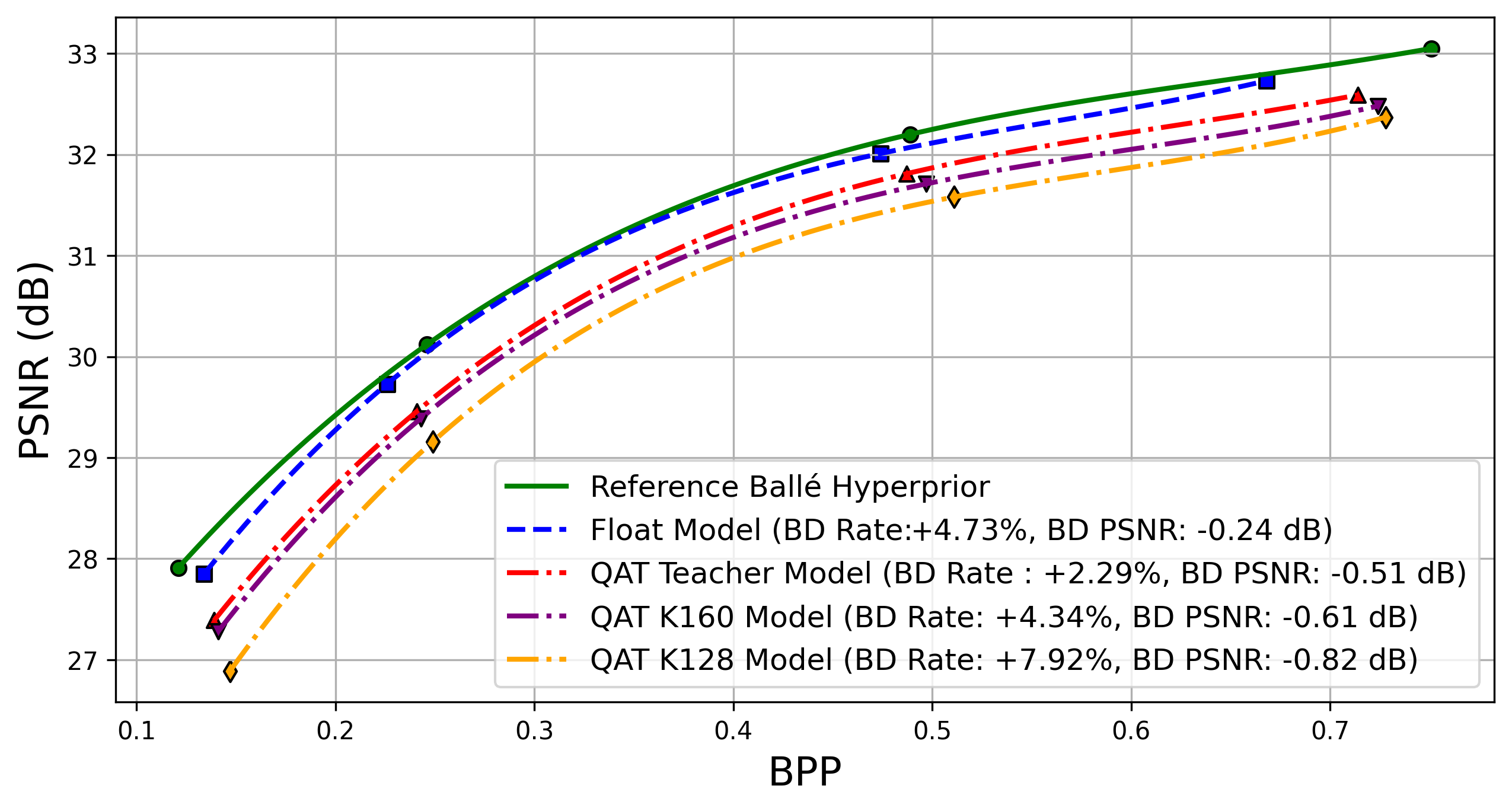}  
  \caption{RD plots an BD-rate, BD-PSNR results using Ballé model as reference on CLIC dataset.}
  \label{fig:clic}
\end{figure}

Fig.\ref{fig:clic} shows the corresponding RD curves on the CLIC dataset \cite{toderici_wenzhe_2020} of the main 192K quantized and pruned hardware model compared to the corresponding float model and the original reference model by Ballé \cite{balle_variational_2018}. The drop in RD from float to hardware is to be expected and is consistent throughout the experiments we have observed, pruning and fixed point quantization result in model degradation, this drop is minimized using fine tuning on both tasks.

Table~\ref{tab:SOTA_energy_fps} presents a comparison of encoding results with the literature. Our proposed implementation achieves the highest frame rate at \textbf{43 FPS}, surpassing the state-of-the-art \textbf{37 FPS}. We attain the fastest hardware-level encoding while using a smaller FPGA board and fewer DSP slices. As a result, our design also achieves the lowest energy consumption per frame at \textbf{2.88 J/frame}, second only to \cite{fpgachen2024}, which does not provide details on the energy measurement process.  

Despite these advancements, \cite{sun2024fpga} achieves a higher DSP utilization efficiency of \textbf{91\%}, compared to our \textbf{85\%}. This difference stems from their exhaustive design space exploration, which optimizes DSP allocation across the pipeline and searched for optimal configurations. However, our approach prioritizes a balanced trade-off between \textbf{throughput, RD performance, and energy efficiency}, making it better suited for real-time encoding scenarios.  

These numbers show how our approach outperforms state-of-the-art competitors like \cite{sun2024fpga} both in RD-efficiency and latency terms and without the need for a complex hardware design space search.
\noindent

\begin{figure}[htbp]
  \centering
  \includegraphics[width=0.8\columnwidth]{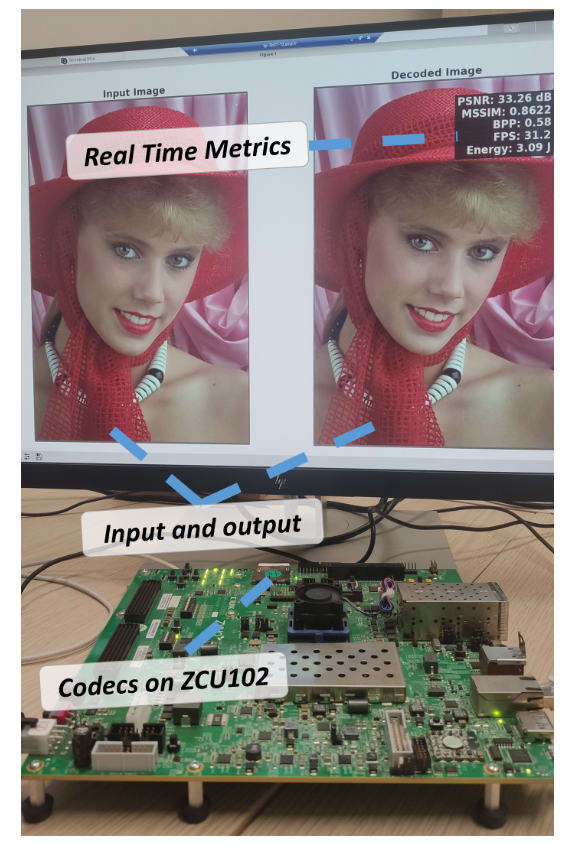}  
  \caption{ZCU102 FPGA setup during experimental validation showing real-time deployment on the embedded Linux}
  \label{fig:place_holder_fpga}
\end{figure}

\begin{table*}[htbp]
\centering
\caption{Hardware System Level Comparisons with the State of the Art}
\small 
\renewcommand{\arraystretch}{1.2} 
\resizebox{\textwidth}{!}{ 
\begin{tabular}{l>{\centering\arraybackslash}p{3 cm}>{\centering\arraybackslash}p{3cm}>{\centering\arraybackslash}p{3cm}>{\centering\arraybackslash}p{3cm}>{\centering\arraybackslash}p{3cm}>{\centering\arraybackslash}p{3cm}} 
\hline
\multicolumn{1}{l}{} & \textbf{Model} &  \textbf{Clock (MHz)} & \textbf{Encoding FPS ↑} & \textbf{DSP Usage} & \textbf{DSP Efficiency ↑} & \textbf{Energy (J/frame) ↓} \\ 
\hline
\textbf{ZCU102 (ours)} & Mid Cost \cite{balle_variational_2018} & 300 & \textbf {43.63} & 2148 & 85\% & 2.88 \\ 
\textbf{ZCU102 \cite{fpgachen2024}} & Low Cost \cite{balle_end--end_2017} & 200 & 32.95 & 1522  & 78\% &\textbf {0.12}  \\ 
\textbf{ZCU104 \cite{jia_fpx-nic_2022}} & High Cost \cite{minnen2018joint} & 300 & 3.90 & NA & NA & 3.51 \\ 
\textbf{KU115 \cite{sun_real-time_2022}} & Mid Cost \cite{balle_variational_2018} & 200 & 25.04 & 2654 & 94\% & 2.95 \\ 
\textbf{VCU118 \cite{sun2024fpga}} & Mid Cost \cite{balle_variational_2018} & NA & 37.82 & 5864 & 93.7\% & NA \\ 
\textbf{KU115 \cite{sun2024fpga}} & Low Cost \cite{balle_end--end_2017} & NA & 43.2 & 3752 & \textbf {93.7\%} & NA \\ 
\hline
\end{tabular}
}
\label{tab:SOTA_energy_fps}
\end{table*}

\subsection{Ablation Study}
In the next section, we perform an ablation study on the Student-160 model, which offers the best RD efficiency while meeting the 25 FPS latency target. We independently remove components including knowledge distillation, GDN implementation, mixed precision quantization, quantization-aware training, and the fully pipelined hardware configuration. For each ablation, we evaluate the impact on RD efficiency, latency, and resource utilization.

\subsubsection{Knowledge Distillation} 
\begin{figure}[htbp]
  \centering
  \includegraphics[width=1\columnwidth]{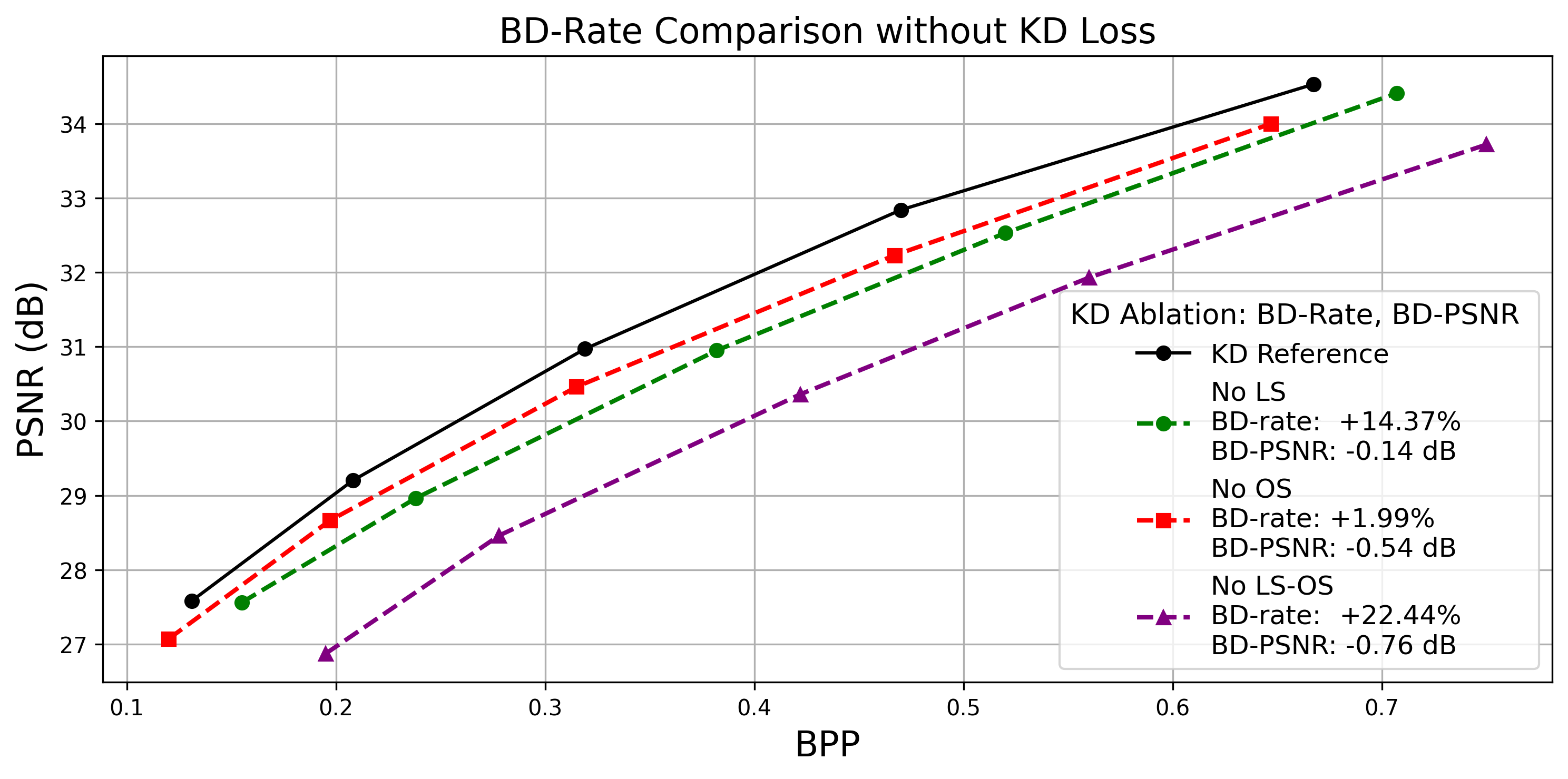}  
  \caption{BD-rate and BD-PSNR results for Latent Space and Output Space ablation showing a 0.76 PSNR drop and 22.4\% rate increase when the KD protocol is not applied
}
  \label{fig:BD_KD}
\end{figure}

We assess the impact of the KD loss terms in Eq.~\ref{eq:kd} on RD-efficiency by fine-tuning the K128 model with the following configurations: 1) both \(\mathcal{L}_{\text{latent}}\) and \(\mathcal{L}_{\text{perc}}\), 2) only \(\mathcal{L}_{\text{latent}}\), 3) only \(\mathcal{L}_{\text{perc}}\), and 4) no loss. Fig.~\ref{fig:BD_KD} shows that omitting the latent space loss increases BD-Rate by 14.37\% and causes a minor drop in PSNR, as the latent loss helps align the student’s learned latent space with the teacher’s, improving compression.

Whereas the output space loss ensures the student’s reconstructed image matches the teacher’s, improving fidelity with lower computational cost. Omitting this loss leads to a 0.54 dB PSNR drop. Removing the entire KD scheme results in a 22.44\% increase in bpp and a 0.76 dB PSNR drop, compared to a +14\% bpp and -0.3 dB drop with KD, as shown in Fig.~\ref{fig:BD_MP}.
This highlights the critical role of both latent and output space KD losses in improving student model performance. The significant BD-rate increase and PSNR drop without these losses emphasize their importance for efficient compression and high-quality reconstruction. These results confirm that KD is essential for closing the performance gap between smaller and larger models, achieving the desired RD and FPS targets.

\begin{figure}[htbp]
  \centering
  \includegraphics[width=1\columnwidth]{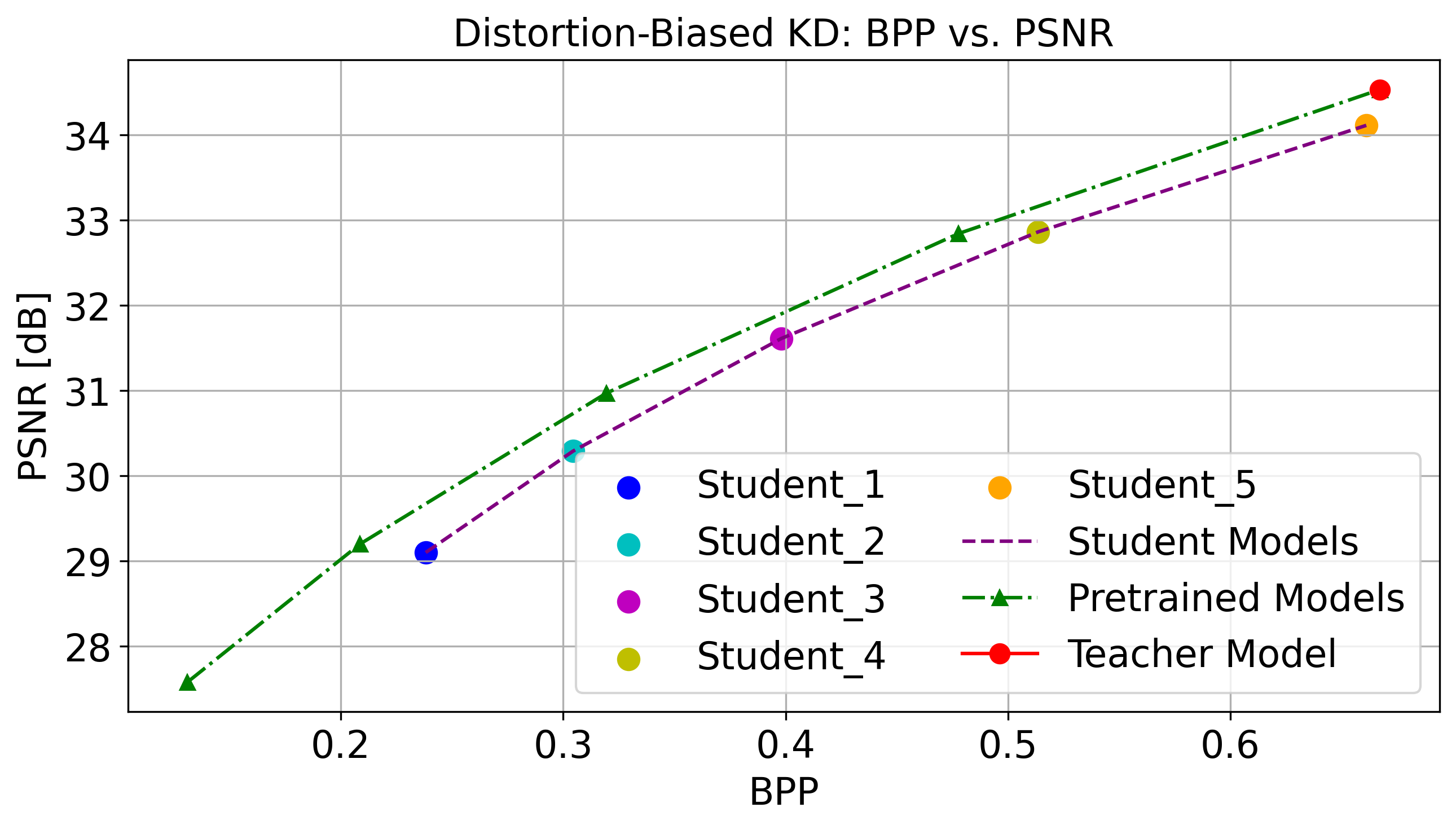}  
  \caption{BD-rate and BD-PSNR results
}
  \label{fig:dist_bias}
\end{figure}

\begin{figure}[htbp]
  \centering
  \includegraphics[width=1\columnwidth]{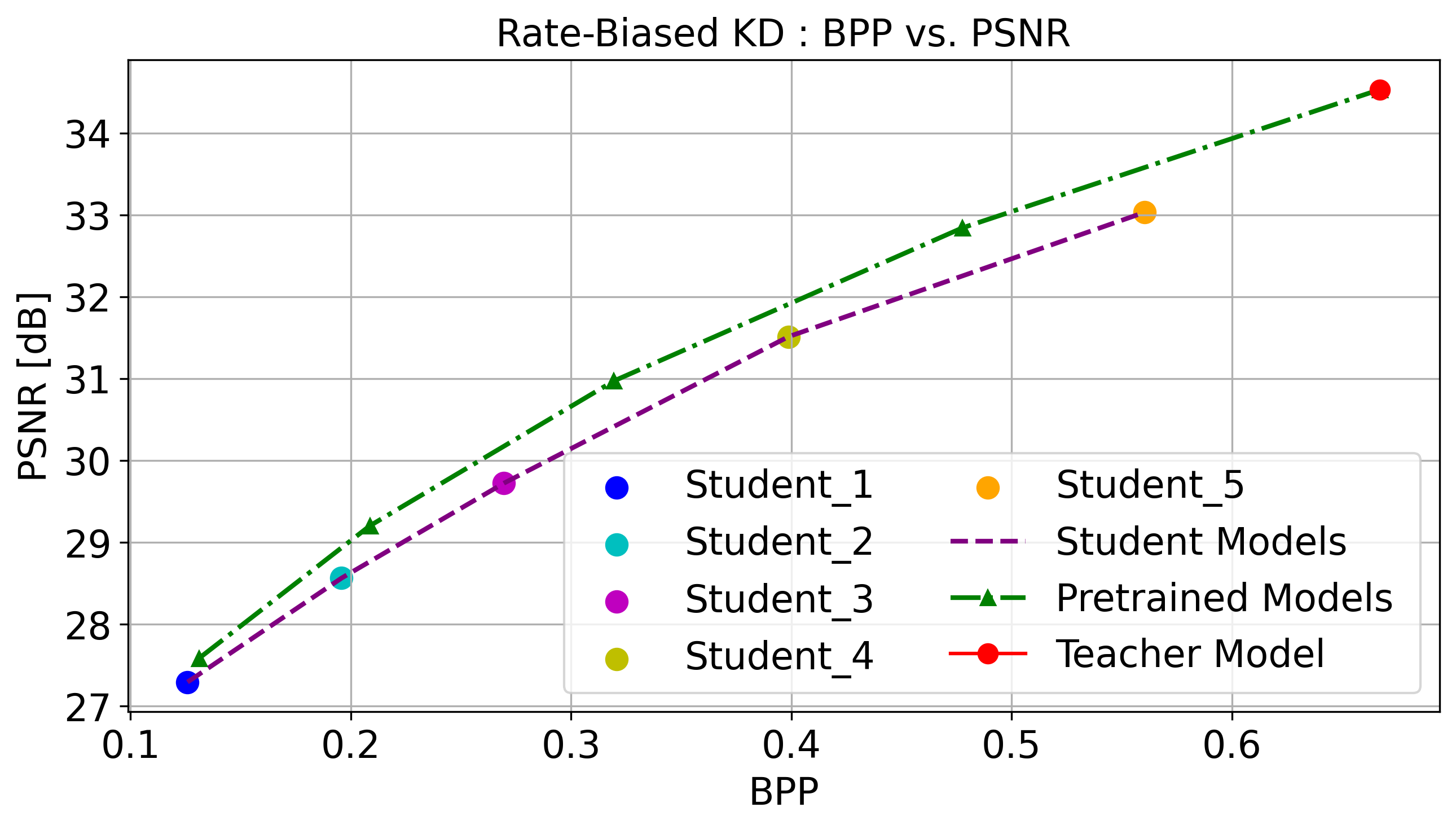}  
  \caption{BD-rate and BD-PSNR results 
}
  \label{fig:rate_bias}
\end{figure}

Now, we investigate the impact of fixing the teacher model during knowledge distillation using two pre-trained teachers from the CompressAI model zoo, differing in \(\lambda\) values: one prioritizing image quality (\(\lambda = 0.0025\), red model in Fig.\ref{fig:dist_bias}) and the other minimizing bit rate (\(\lambda = 0.18\), red model in Fig.\ref{fig:rate_bias}). In previous experiments, we always use the same \(\lambda\) for the Teacher-Student pair.  The students, each with half the number of filters of the teachers, are trained with these two different teachers to analyze how the teacher’s focus affects the students' RD performance.
When the teacher emphasizes image quality, smaller student models benefit from higher PSNR than their pre-trained \(\lambda\) counterparts, despite having fewer channels. However, this comes at the expense of increased bit rate. As \(\lambda\) decreases, the effect becomes less pronounced, with the student models showing a +8.75\% BD-rate and -0.39 dB BD-PSNR compared to the baseline teacher. In contrast, when the teacher focuses on minimizing bit rate, the students are pushed towards lower bit rates but suffer a reduction in image quality. This trade-off is more pronounced as \(\lambda\) increases, resulting in a +10.32\% BD-rate and -0.42 dB BD-PSNR compared to the baseline teacher.

While knowledge distillation benefits smaller models over training from scratch, its effectiveness depends on both the teacher’s initial performance, "priorities" and the student model size. Students trained with an distortion-focused teacher, seen in Fig.\ref{fig:dist_bias} show improved PSNR but still cannot match the teacher's RD performance due to the trade-off in rate. On the other hand, students trained with a rate-focused teacher, Fig.\ref{fig:rate_bias}, achieve lower bit rates at the cost of image quality. This experiment highlights that the benefits of KD are shaped by the teacher's priorities and the student model’s size, particularly when balancing rate and distortion and the possibility of using a teacher-student pair with varying \(\lambda\) values based on the objective.

\subsubsection{GDN vs ReLU}

\begin{figure}[htbp]
  \centering
  \includegraphics[width=1\columnwidth]{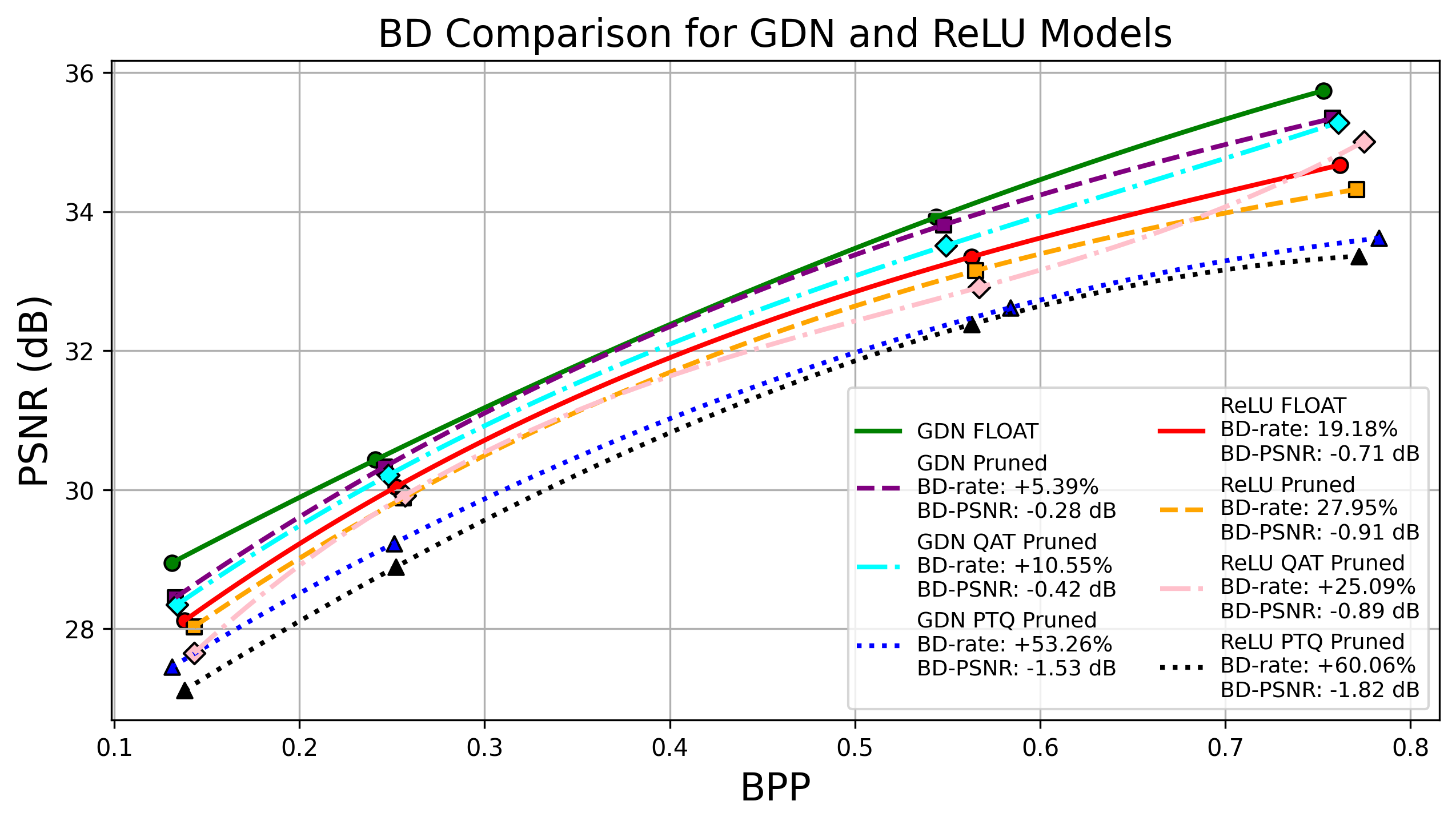}  
  \caption{BD-Rate and BD-PSNR results for several model permutations with a float model using GDN as reference
}
  \label{fig:GDN_comparison_mp}
\end{figure}

Next, we drop our GDN implementation in favor of a simpler ReLU activation, as in \cite{jia_fpx-nic_2022,sun2024fpga, sun_real-time_2022}.
While a drop in RD efficiency is expected~\cite{balle_integer_2018}, our work is the first to quantitatively assess it on an FPGA platform as we are the first to propose a GDN implementation on FPGA.
Fig. \ref{fig:GDN_comparison_mp} provides a better look into the difference in performance when using GDN instead of ReLU, in general an average drop of between  -0.5 and -0.7 dB BD-PSNR and increase +10\% and 25\% BD-rate is observed throughout the different permutations, this goes in line with the insights gleaned in \cite{balle_integer_2018}. Therefore our GDN implementation pushed the RD-efficiency of our method by a significant margin giving our hardware LIC model the edge over the state of the art \cite{sun2024fpga}.  

\subsubsection{Quantization strategy} 

\begin{figure}[htbp]
  \centering
  \includegraphics[width=1\columnwidth]{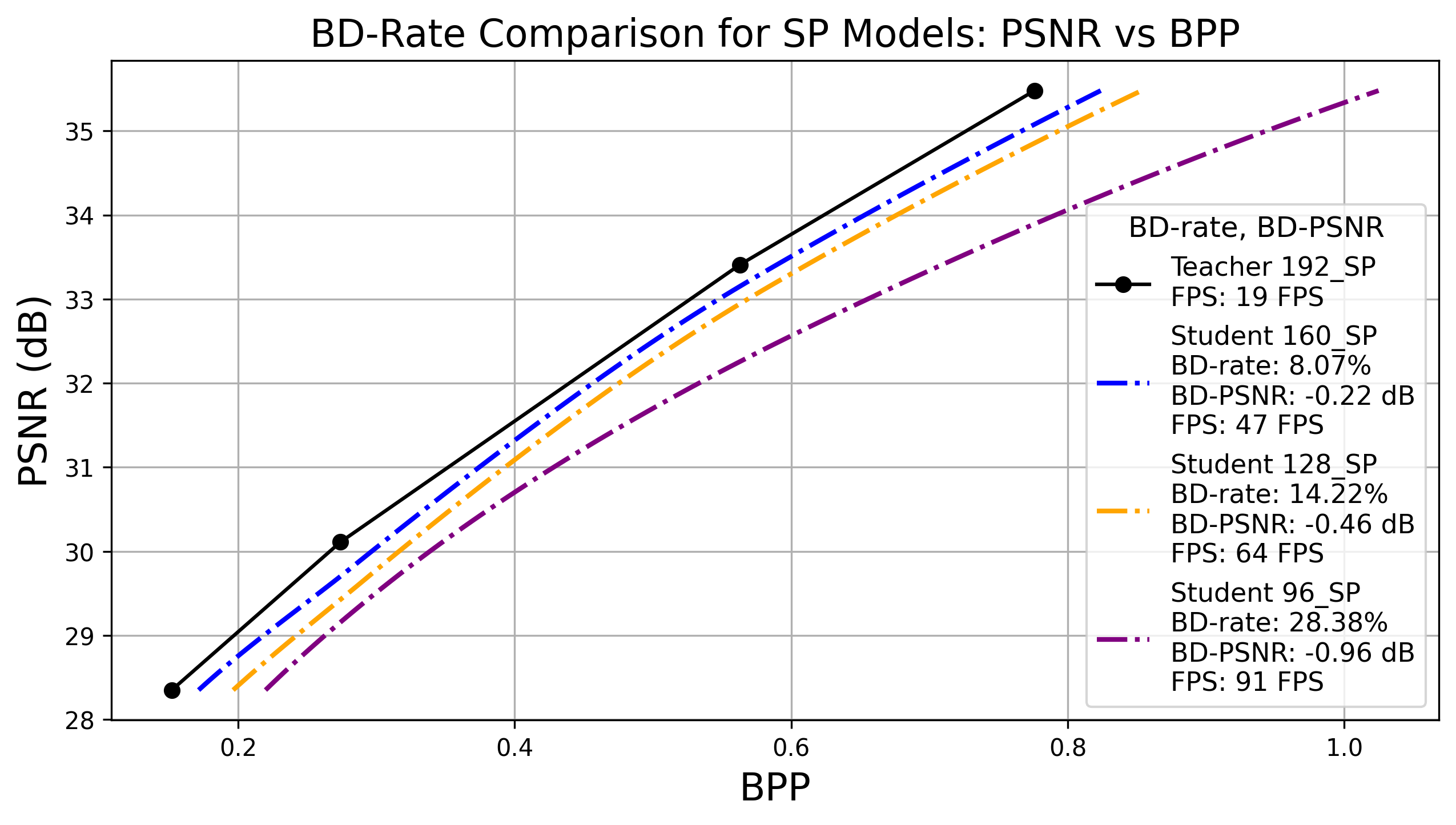}  
  \caption{BD rate and PSNR for Single Precision models}
  \label{fig:BD_SD}
\end{figure}

\begin{figure}[htbp]
  \centering
  \includegraphics[width=1\columnwidth]{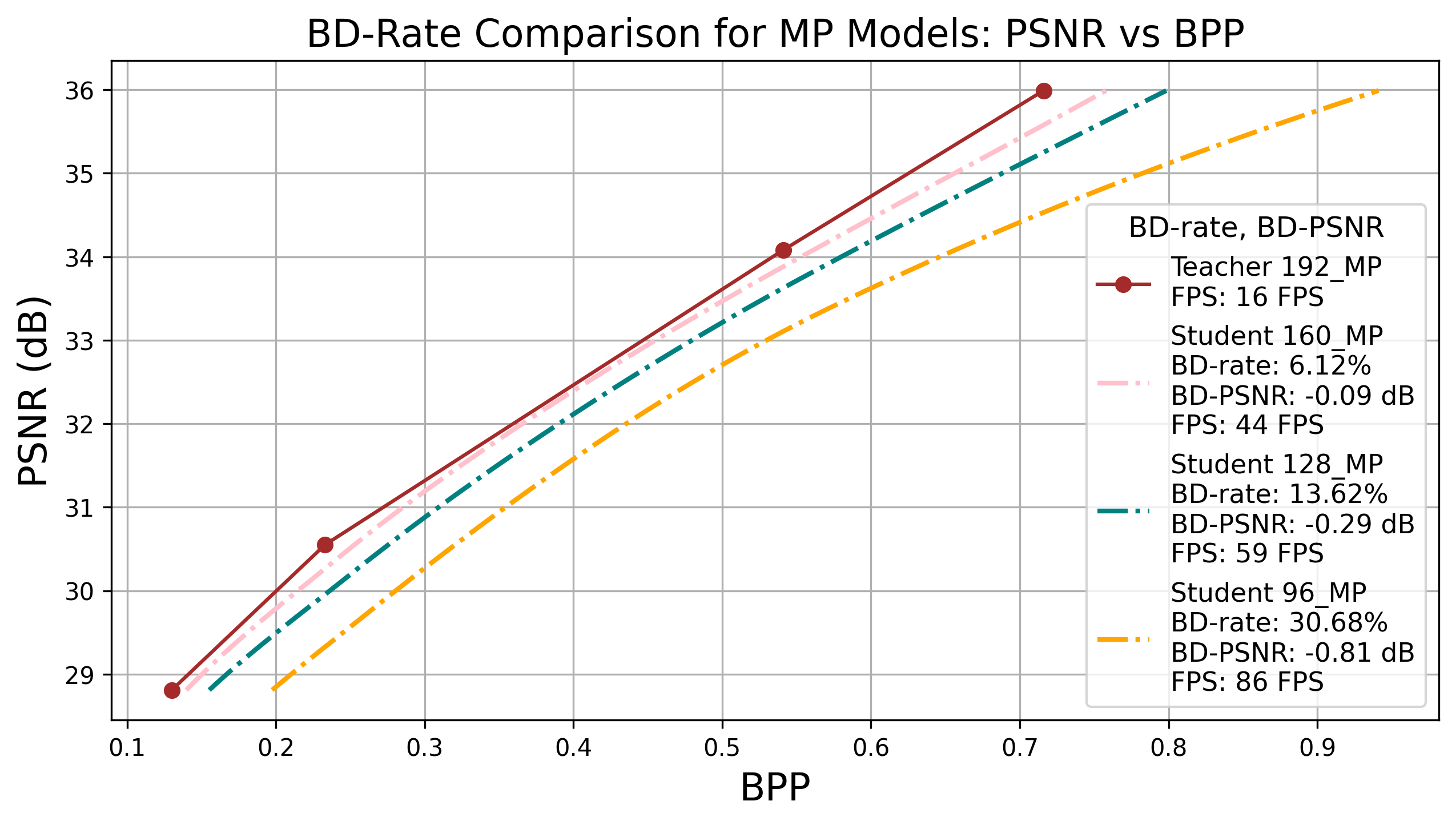}  
  \caption{BD rate and PSNR for Mixed Precision models}
  \label{fig:BD_MP}
\end{figure}

As a third set of ablation experiments, we experiment with two different aspects of the quantization strategy.
\\
{a) Mixed Precision vs Single Precision:} 
As a first experiment, instead of allocating 32 rather than 8 bits to the GDN parameters to boost its precision (mixed precision), we allocate 8 bits to all layers, i.e. we implement a single precision quantization scheme.
Assigning higher precision to the quantization-sensitive GDN activation layers consistently improves PSNR, averaging +0.3 dB across all model configurations. Fig.~\ref{fig:BD_MP} and \ref{fig:BD_SD} illustrate the benefits of hybrid quantization, with a +0.2 dB PSNR improvement and -2\% BPP. The improvement is less significant for smaller student models, likely due to their limited ability to leverage the higher precision representation, given their fewer input and output channels per layer.

{b) QAT vs. PTQ Quantization:} Next, we replace Quantization Aware Training where quantization is jointly carried out with a fine-tuning step for a simpler Post Training Quantization (PTQ) scheme where the model is quantized post-training without any further fine-tuning.
Fig.~\ref{fig:GDN_comparison_mp} shows the RD curves for both quantization schemes. Compared to QAT, PTQ results in a significant RD efficiency drop, with a 40\% increase in BD-rate and a -1 dB decrease in BD-PSNR. This experiment highlights the advantages of integrating the quantization step into the learning loop when minimizing a multi-loss function during training.
\begin{table*}[htbp]
\centering
\caption{Single Precision KD Distillation Results on Kodak 720p Inputs}
\renewcommand{\arraystretch}{1.2} 
\resizebox{\textwidth}{!}{ 
\begin{tabular}{l>{\centering\arraybackslash}p{2cm}>{\centering\arraybackslash}p{2cm}>{\centering\arraybackslash}p{2cm}>{\centering\arraybackslash}p{2cm}>{\centering\arraybackslash}p{2cm}>{\centering\arraybackslash}p{2cm}>{\centering\arraybackslash}p{2cm}>{\centering\arraybackslash}p{2cm}} 
\hline
\multirow{2}{*}{\text{Lambda}} & \multicolumn{2}{c}{\textbf{Teacher K192}} & \multicolumn{2}{c}{\textbf{Student K160}} & \multicolumn{2}{c}{\textbf{Student K128}} & \multicolumn{2}{c}{\textbf{Student K96}} \\
\cline{2-9}
                                   & \textbf{Rate [bpp] ↓} & \textbf{PSNR [dB] ↑} & \textbf{Rate [bpp] ↓} & \textbf{PSNR [dB] ↑} & \textbf{Rate [bpp] ↓} & \textbf{PSNR [dB] ↑} & \textbf{Rate [bpp] ↓} & \textbf{PSNR [dB] ↑} \\
\hline
\text{0.0016}                    & 0.152      & 28.35     & 0.136      & 28.22     & 0.144      & 27.66     & 0.151      & 27.29     \\
\text{0.0032}                    & 0.274      & 30.21     & 0.243      & 29.82     & 0.262      & 29.67     & 0.287      & 28.77     \\
\text{0.0075}                    & 0.563      & 33.41     & 0.573      & 33.35     & 0.583      & 33.14     & 0.612      & 32.66     \\
\text{0.0450}                     & 0.776      & 35.48     & 0.758      & 35.18     & 0.786      & 35.12     & 0.797      & 34.11     \\
\hline
\end{tabular}%
}
\label{tab:BD_SP}
\end{table*}

\begin{table*}[htbp]
\centering
\caption{Mixed Precision KD Distillation Results on Kodak 720p Inputs}
\renewcommand{\arraystretch}{1.2} 
\resizebox{\textwidth}{!}{ 
\begin{tabular}{l>{\centering\arraybackslash}p{2cm}>{\centering\arraybackslash}p{2cm}>{\centering\arraybackslash}p{2cm}>{\centering\arraybackslash}p{2cm}>{\centering\arraybackslash}p{2cm}>{\centering\arraybackslash}p{2cm}>{\centering\arraybackslash}p{2cm}>{\centering\arraybackslash}p{2cm}} 
\hline
\multirow{2}{*}{\text{Lambda}} & \multicolumn{2}{c}{\textbf{Teacher K192}} & \multicolumn{2}{c}{\textbf{Student K160}} & \multicolumn{2}{c}{\textbf{Student K128}} & \multicolumn{2}{c}{\textbf{Student K96}} \\
\cline{2-9}
                                   & \textbf{Rate [bpp] ↓} & \textbf{PSNR [dB] ↑} & \textbf{Rate [bpp] ↓} & \textbf{PSNR [dB] ↑} & \textbf{Rate [bpp] ↓} & \textbf{PSNR [dB] ↑} & \textbf{Rate [bpp] ↓} & \textbf{PSNR [dB] ↑} \\
\hline
\text{0.0016}                    & 0.142      & 28.61     & 0.134      & 28.52     & 0.137      & 28.31     & 0.148      & 27.77     \\
\text{0.0032}                    & 0.247      & 30.25     & 0.223      & 30.18     & 0.232      & 29.98     & 0.263      & 29.48     \\
\text{0.0075}                    & 0.551      & 33.92     & 0.531      & 33.88     & 0.563      & 33.79     & 0.603      & 33.22     \\
\text{0.0450}                     & 0.763      & 35.83     & 0.724      & 35.78     & 0.766      & 35.68     & 0.785      & 34.73     \\
\hline
\end{tabular}%
}
\label{tab:BD_MP}
\end{table*}

\subsubsection{Sequential vs Pipelined:}

Next, we assess the impact of our fully pipelined implementation on latency compared to a simpler sequential configuration. Our design, shown in Fig.~\ref{fig:pipe}, enables parallel operations that fully exploit the three DPU cores. In contrast, other implementations in the literature \cite{jia_fpx-nic_2022} use a unified computational lane, leading to bandwidth and resource inefficiencies due to idle resources. We therefore test a baseline sequential configuration where all components (main encoder, hyperprior encoder, entropy encoder-decoder, hyperprior decoder, and main decoder) are executed consecutively. Although each component utilizes all DPU cores, this approach results in suboptimal resource usage during intermediate stages.
A unified computational lane is dedicated to all processes, with each module processed sequentially and intermediate results stored in external memory as needed. Table~\ref{tab:seq_vs_pipe_160} presents the performance of the K60 teacher model under both configurations. The pipelined design achieves 2.5× higher FPS thanks to improved DSP efficiency and external memory bandwidth. The distilled 160K student model shows significant gains, with knowledge distillation reducing model size and FLOPs, yielding 10.8 GB/s bandwidth and 87\% DSP efficiency. The student model reaches 42.6 FPS in the pipelined encoder—nearly 3× faster than the sequential version. Overall, the pipelined design fully exploits the ZCU102 FPGA, offering substantial throughput benefits.

\begin{table}[htbp]
\centering
\scriptsize 
\renewcommand{\arraystretch}{1.0} 
\caption{Model latency as a function of the implementation configuration for a single 720p image on the 160K student.}
\begin{tabular}{l c c c c c c}
\hline
 & \multicolumn{3}{c}{\textbf{Sequential}} & \multicolumn{3}{c}{\textbf{Pipelined}} \\ 
\cline{2-7}
 & \textbf{FPS $\uparrow$} & \textbf{DSP\% $\uparrow$} & \textbf{BW} & \textbf{FPS $\uparrow$} & \textbf{DSP\% $\uparrow$} & \textbf{BW} \\
\hline
\textbf{Encoder} & 17.3  & 56\% & 3.5 GB/s & 43.6  & 87\% & 7.1 GB/s \\
\textbf{Decoder} & 13.6   & 51\% & 3.3 GB/s & 30.5  & 85\% & 6.7 GB/s \\
\hline
\end{tabular}
\label{tab:seq_vs_pipe_160}
\end{table}

\section{Conclusion and future works}
This work presents a hardware-efficient FPGA implementation of a Learned Image Compression (LIC) model, achieving real-time latency and high RD efficiency. A novel Knowledge Distillation-based transfer learning scheme reduces computational cost while maintaining performance, using larger models as teachers to guide smaller models in jointly learning better reconstruction and rate representations. 
We introduce an FPGA-optimized Generalized Divisive Normalization (GDN) layer with a mixed-precision approach, yielding PSNR gains and bit-rate reduction while minimizing FLOPS. 
Optimizations through Quantization-Aware Training (QAT) and structured pruning reduce complexity, enhancing FPGA efficiency without sacrificing quality. The design employs patch-parallel pipelining, achieving real-time 720p performance and outperforming existing RD metrics with up to three times the performance of sequential designs.
Future work will extend knowledge distillation by exploring alternative loss functions, including one for activation maps to better guide the student. We will also investigate freezing the hyperprior model, extending KD to the hyperprior, or focusing on distilling the decoder for edge applications. Further refinements in dynamic weighting and continued software-hardware co-modeling will optimize mapping configurations.

\bibliographystyle{IEEEtran}  
\bibliography{sampleBibFile}  
\end{document}